\documentclass[11pt]{article}
\usepackage[margin=1in]{geometry}
\usepackage{lmodern}
\usepackage[T1]{fontenc}
\usepackage{amsmath,amssymb}
\usepackage{booktabs}
\usepackage{graphicx}
\usepackage{xcolor}
\usepackage[hidelinks]{hyperref}
\usepackage{caption}
\usepackage{microtype}
\graphicspath{{plots/}}

\usepackage{tikz}
\usetikzlibrary{positioning,calc,arrows.meta}
\usepackage{array}
\usepackage{longtable}
\usepackage{float}
\usepackage{placeins}
\definecolor{vok}{HTML}{3B6D11}
\definecolor{vno}{HTML}{A32D2D}
\definecolor{vwk}{HTML}{854F0B}
\definecolor{vna}{HTML}{888780}
\definecolor{vdet}{HTML}{5F5E5A}
\definecolor{bodyfill}{HTML}{EEEDFE}\definecolor{bodydraw}{HTML}{534AB7}\definecolor{bodytext}{HTML}{26215C}
\definecolor{vocfill}{HTML}{FAECE7}\definecolor{vocdraw}{HTML}{993C1D}\definecolor{voctext}{HTML}{4A1B0C}
\definecolor{nfill}{HTML}{F1EFE8}\definecolor{ndraw}{HTML}{5F5E5A}
\newcommand{\vyes}[1]{{\color{vok}\ensuremath{\checkmark}}~{\scriptsize\color{vdet}#1}}
\newcommand{\vnox}[1]{{\color{vno}\ensuremath{\times}}~{\scriptsize\color{vdet}#1}}
\newcommand{\vwkk}[1]{{\color{vwk}\ensuremath{\sim}}~{\scriptsize\color{vdet}#1}}
\newcommand{\vnaa}{{\color{vna}--}}
\newcommand{\lk}[1]{\textcircled{\scriptsize#1}}
\newcommand{\drawspine}[1]{%
  \node[box]  (teacher)  at (0,0)        {Teacher trait / policy\\{\scriptsize fine-tune $\to\Delta\theta_T$}};
  \node[box]  (signal)   at (3.4,0)      {Distillation signal\\{\scriptsize teacher logits; $\tau$/markers masked}};
  \node[box]  (student)  at (6.8,0)      {Student update\\{\scriptsize initial $d_0$, then body movement}};
  \node[bbox] (body)     at (10.2,1.15)  {Body computation\\{\scriptsize behaviour carrier}};
  \node[vbox] (vocab)    at (10.2,-1.15) {Unembedding readout\\{\scriptsize convergent geometry}};
  \node[box]  (transfer) at (13.6,0)     {Transfer\\{\scriptsize $P(\tau)$ or interaction $I$}};
  \draw[ar] (teacher.east)  -- (signal.west);
  \draw[ar] (signal.east)   -- (student.west);
  \draw[ar] (student.east)  -- (body.west);
  \draw[ar] (student.east)  -- (vocab.west);
  \draw[ar] (body.east)     -- (transfer.west);
  \draw[ar] (vocab.east)    -- (transfer.west);
  \draw[dashed,draw=ndraw,line width=0.4pt] (teacher.north) to[bend left=30]
      node[badge,pos=0.5]{3}
      node[pos=0.5,above=5mm,font=\scriptsize,text=vdet]{#1} (student.north);
  \node[badge] at (1.7,0)                   {1};
  \node[badge] at (5.1,0)                   {2};
  \node[badge] at (8.4,0)                   {4};
  \node[badge] at ([yshift=3mm]vocab.north) {5};
  \node[badge] at (11.95,0)                 {6};
}
\tikzset{
  box/.style={draw=ndraw,fill=nfill,rounded corners=3pt,align=center,
              text width=2.5cm,inner sep=3pt,font=\footnotesize,minimum height=0.9cm},
  bbox/.style={box,draw=bodydraw,fill=bodyfill,text=bodytext},
  vbox/.style={box,draw=vocdraw,fill=vocfill,text=voctext},
  ar/.style={-{Stealth[length=2mm]},draw=ndraw,line width=0.4pt},
  badge/.style={circle,draw=black!45,fill=white,inner sep=0pt,minimum size=4.6mm,font=\scriptsize},
}
\title{\textbf{Channel Location Constrains the Auditability of Subliminal Learning}}
\author{Tamas Madl\\[2pt]
  Austrian Research Institute for Artificial Intelligence\\
  University of Vienna\\[2pt]
  \texttt{tamas.madl@ofai.at}\\
  \href{https://orcid.org/0000-0001-6120-7855}{orcid.org/0000-0001-6120-7855}}
\date{}

\begin{document}
\maketitle

\begin{abstract}
Subliminal learning lets a student inherit a teacher's hidden trait from distillation data that never names it. We ask when such transfer can be audited before training. The answer is not model identity or scale alone, but channel location: the carrier through which the trait reaches the student. We find three regimes. In a controlled initialization-dependent body channel, a pre-training screen works. Coverage, the cosine between the student's initial distillation update and the teacher's fine-tuning displacement, predicts held-out transfer (Spearman $\rho \approx 0.95$; AUROC 0.997). In pretrained language models, masked single-token traits instead ride convergent vocabulary geometry. This channel is initialization-independent, so initialization-alignment screens, including coverage, are not mechanistic; the useful handles are post-hoc detection and targeted mitigation. Even when a single-token named entity is removed from the loss, the student's held-out probability for that entity rises to 0.40 on average ($\sim 2500\times$), and a related semantic class transfers. In an untied-head model, orthogonalizing the trait's output row against entangled neighbours collapses leakage, while equal-size random-subspace edits do not. Thus removing a target string from distillation labels does not remove the corresponding preference: neighbouring tokens can carry it. Finally, conditional behaviours can route through the network body. For sycophancy, with agreement and correction markers masked from the loss, transfer reaches about 0.63 of the teacher's effect, localizes to body computation, and evades four audits across two model families. We scope this as masked transfer of a condition-present policy. Channel location is necessary for deciding which audits can be sound. It is not a deployment-ready screen: an audit used outside its carrier regime can give false assurance.
\end{abstract}

\section{Introduction}

Cloud et al.\ \cite{cloud2025subliminal} showed that a student model fine-tuned on data generated by a teacher with some
trait (a preference, a persona, a misalignment) can acquire that trait even when the data are
number sequences or noise with every explicit reference to the trait removed, provided teacher and
student share an initialization. The effect is a genuine safety concern for standard teacher--student
distillation pipelines, in which a student is trained to match a teacher's outputs
\cite{hinton2015distilling,stanton2021does}: a trait can ride through ostensibly neutral data.
Their analysis is mechanistic but establishes only a binary condition: under a shared initialization,
a gradient step on any teacher output either moves the student toward the teacher or is orthogonal
to it. It thus settles whether transfer can happen, but not how much.

Consider a concrete deployment failure. A distilled assistant is trained on an untrusted teacher's
outputs after every occurrence of a sensitive target (a brand, organization, political figure, product
label, or other named entity) has been removed from the distillation loss. A conventional audit of
the supervised targets passes: the forbidden string is never trained on directly. But that audit checks
the wrong boundary. The teacher can shift probability mass onto high-cosine output-row neighbours,
often semantically related ones, and the student can inherit the target preference through that
residue. This is shown in our token-trait experiments. With a single-token named commercial entity
masked from the loss, the student's held-out probability for it still rises to $0.40$ on average
(a mean per-entity lift of $\sim2500\times$), and a preference for a related \emph{class} of entities
transfers likewise. The carrier is token-geometric rather than entity-semantic: the object whose
removal stops the leakage is the entity token's unembedding-neighbour structure, not the real-world
entity it denotes. We find no carrier-level distinction between these low-harm proxies and higher-risk
entity classes, which we therefore do not instantiate. So ``the target string is absent from the
training labels'' does not imply ``the target preference is absent from the trained model'': filtering
the token did not filter the preference. This belongs to the same family of deployment failure that
recent data-level defence results expose from another angle: removing, filtering, or paraphrasing
suspicious surface form is not a reliable safety boundary when the carrier is distributed across the
teacher's output distribution \cite{behrens2025dataset,draganov2026phantom,gisler2026faithful}.
This is the latent steering bias to detect before deploying a distilled model,
especially when the teacher is external, proprietary, or trained under unknown incentives.

This paper asks whether the amount of transfer can be computed before the student is trained,
turning ``did it transmit?'' from a post-hoc audit into a prospective pre-training audit. We call any
quantity computed before training that aims to predict transfer a \emph{screen}, and the part of the
channel whose removal, replacement, or neutralization abolishes transfer the channel's \emph{carrier}.
Whether a screen is even possible depends on where the carrier lives. A carrier in the
initialization-dependent network body, meaning the non-output weights, can be screenable; a carrier
in convergent \emph{unembedding geometry}---the output matrix mapping hidden states to vocabulary logits, which we also call the vocabulary or readout geometry---is not screenable by initialization-alignment tests; and a
single pretrained model can hide a carrier in either place. Channel location---where the hidden-transfer channel's carrier sits---rather than model identity or scale alone, constrains auditability
(Table~\ref{tab:audit-map}).

\begin{table}[h]
\centering\footnotesize
\setlength{\tabcolsep}{4pt}
\begin{tabular}{@{}llcccc@{}}
\toprule
Regime & Carrier & Init.\ dep. & Pre-train screen & Post-hoc audit & Intervention \\
\midrule
Toy auxiliary trait & body displacement & yes & works ($\rho\,0.95$) & --- & channel-dep. \\
LM token trait & convergent vocab.\ geom. & no & blind (non-mech.) & elevated-token scan & $W_\tau$ ablation \\
Conditional behaviour & body computation & partial & weak${}^{*}$ & no token handle & open problem \\
\bottomrule
\end{tabular}
\caption{\textbf{Channel location, not model identity, determines which class of audit can be valid.}
Each row is a regime. The screen column reports whether the initialization-alignment screen
(coverage) works, is non-mechanistic and non-deployable, or is only weakly positive
($^\ast$: directional but not thresholdable, because the body-carried behaviour's first update has
only a minority component along the teacher direction; Section~\ref{sec:behavior}). Trusting a screen in the wrong
regime creates false reassurance: a low coverage score or a clean token scan can read ``safe'' on a
channel that still transfers. The right-hand column gives the best audit or intervention we find for
each regime; the token-trait row is refined in Table~\ref{tab:cells}. Coverage is one cell of this map, not the map
itself.}
\label{tab:audit-map}
\end{table}

This paper does not propose a new universal predictor of subliminal transfer; it develops a
channel-conditioned auditability map that organizes, rather than competes with, the recent plurality
of mechanism proposals. The claim is conditional, not universal: where the carrier resides determines
which audit stage can be sound. Coverage, an initial gradient--displacement alignment score, is the
positive case---predictive, mechanistic, and prospectively useful where the carrier is an
initialization-dependent body channel, and non-deployable once the carrier moves to convergent
unembedding geometry or to weakly aligned body computation.

Sections~\ref{sec:law}--\ref{sec:safety} establish this taxonomy across
its three regimes and fill in its cells. Section~\ref{sec:law} gives a positive result: in a controlled auxiliary-channel
setting the carrier is in the network body, and coverage---a steering-angle check computed from the
initialization alone, asking whether the first update induced by the distillation data points the way
the teacher actually moved---predicts transfer. We establish this with a powered sweep, a held-out
prospective trial, a rival-predictor benchmark, a dose-response intervention, and an out-of-distribution
replication. Section~\ref{sec:realLM} gives the contrasting negative result: in pretrained language models a
single-token trait is instead carried by convergent unembedding geometry. It therefore transfers initialization-independently, so any
initialization-alignment screen, coverage included, is non-mechanistic and non-deployable for that
channel, and the effect does not fade across the scales we test. Section~\ref{sec:behavior} shows that the same
pretrained model can instead route a conditional behaviour through the network body, making
channel location a property of the signal rather than the model; a relocation experiment adds
directional support, restoring initialization-gating once the readout is a random body channel.
Section~\ref{sec:safety} evaluates the audit lifecycle---pre-training screening, post-hoc detection, and mitigation---and
shows where each regime can and cannot be caught. Our claims span an evidence gradient; we label
the evidence status of each headline claim explicitly in Table~\ref{tab:claims}---a conventionally powered toy law,
replicated causal dissociations, and safe-proxy demonstrations---and flag which is which throughout.
A screen trusted outside its regime is unsound.

The deployment conclusion is mixed: a strong pre-training screen exists in
the controlled body-channel setting, but no validated pre-training screen exists for any naturally
occurring channel we test. What survives is a causal intervention---targeted output-row editing for
vocabulary-carried token traits, which are also post-hoc detectable---and teacher- and
distillation-pipeline governance for body-carried behaviours, which evade every audit stage we
tested across two model families. This regime is the closest point of contact with the
security literature on persistent trigger-conditioned policies and backdoor transfer by distillation:
the dangerous analogue is not a token preference with an obvious output handle, but a conditional
computation whose trigger and policy may remain dormant under ordinary post-hoc scans \cite{hubinger2024sleeper,cheng2024transferring}.

Recent work does not point to a single universal carrier for subliminal transfer: depending on the
construction, it emphasizes shared-initialization or gradient alignment, sparse divergence tokens and
early layers, compatible output heads, entangled vocabulary rows, steering-vector distillation, or
log-linear data selection \cite{cloud2025subliminal,schrodi2025towards,zur2025token,adenali2026subliminal,brockers2026noise,okatan2025seed,kitkana2026sustained,blank2026steering}.
These are not mutually exclusive accounts: each is, in effect, a claim about which carrier a given construction activates. Our claim is
that the carrier is construction- and signal-dependent, and that this dependence is operational:
channel location determines which audit stage can be sound---pre-training screening, post-hoc
detection, or targeted mitigation. Full related work is in Section~\ref{sec:related}.

\paragraph{Three senses of a valid screen.}
A pre-training screen can be valid in three senses, which our results show come apart.
(1) \emph{Predictive}: its score correlates with transfer across traits sampled from one model.
(2) \emph{Mechanistic}: it reads a quantity the channel actually depends on, so that intervening on
the score---or changing the initialization---moves transfer with it. (3) \emph{Deployable}: a fixed
threshold delivers a stated false-negative rate on an unknown trait, possibly in a different model.
Only a mechanistic screen can be deployable, and only a deployable screen can be trusted in the
field. Coverage is predictive and mechanistic in the controlled body-channel setting; in the
real-model body-computed behaviours we test, the same quantity is only a weak directional symptom,
not a thresholdable screen. For vocabulary-carried traits it is merely predictive: within one family its
scores even correlate with transfer, but the channel is initialization-independent, so that correlation
is inert---unmoved by intervention. On tied Qwen it is not even predictive, remaining at chance.
When we call a screen blind to a channel we mean this operational sense---non-mechanistic and
non-deployable---whether or not it happens to correlate within a single family. We use this
trichotomy, framed as the audit decision problem in Section~\ref{sec:safety}, to state precisely what each regime's
screen can and cannot be trusted to do, rather than reporting a single correlation.

\paragraph{Contributions.}
\textbf{(i) A channel-conditioned auditability map.}
Subliminal transfer is not governed by a single universal carrier or a single universal audit: the valid
audit stage depends on where the carrier resides. One structural variable organizes all three regimes:
whether the channel lives in the initialization-dependent network body or in convergent
unembedding geometry. An initialization-dependent body channel can admit a pre-training
alignment screen; a convergent output-readout channel evades such screens but is detectable and
mitigable post hoc; and body-computed conditional behaviours remain the open audit gap. The
carrier assignments are established by orthogonalization ablations, neighbour interventions, and
body/head transplants, and a relocation experiment adds convergent directional support: moving a
trait into a random body channel restores initialization-gating and coverage's ordering near the
transfer floor (Section~\ref{sec:behavior}).

\textbf{(ii) A powered screen for the body-channel regime.}
Coverage, $\cos(d_0,\hat{u}_T)$---the cosine between the student's initial distillation update $d_0$
and the unit teacher displacement $\hat{u}_T$, computable at the shared initialization with no
student training---predicts held-out subliminal-transfer accuracy in a controlled body channel,
outperforms every inexpensive rival, acts as a causal lever, and predicts trait specificity (which
teacher's trait a student copies) and cross-initialization failure (Section~\ref{sec:law}). A channel-Fisher identity
makes coverage and specificity the normalized diagonal and off-diagonal of one bilinear form
(Section~\ref{sec:identity}). It is predictive, mechanistic, and prospectively useful in the controlled body-channel
regime: the positive case showing what a sound screen looks like, not a universal defence. Gradient alignment as the body-channel mechanism is also identified by Kitkana \& Arora \cite{kitkana2026sustained}; our contribution
is its calibrated, a-priori form.

\textbf{(iii) A causal localization of vocabulary-carried token transfer.}
In pretrained language models, we localize single-token subliminal transfer to a readout dependence
on unembedding entanglement: high-cosine coupling among output rows induced by the softmax
bottleneck when the vocabulary is much larger than the hidden width. In the removal-test sense, the
body supplies the displacement, but the entangled output geometry is required for expressing the
masked trait. Orthogonalizing the target row against its entangled neighbours drives leakage to the
floor while preserving supervised transfer and perplexity, and an equal-dimension random-subspace
placebo leaves leakage intact. The localized channel does not fade over the scales we test: it rises
with scale in Pythia and is at or near ceiling in the other families, up to nearly seven billion
parameters in full precision. This pattern is inconsistent with a simple bottleneck-ratio fade
hypothesis, although cross-family differences in magnitude are capability-confounded
(Sections~\ref{sec:realLM} and~\ref{sec:safety}).

\textbf{(iv) A body-mediated conditional behaviour.}
A conditional behaviour---sycophancy, measured as a false-vs-true agreement interaction---also
transfers subliminally, but localizes predominantly to the network body rather than the vocabulary.
It dissociates from the token traits on the neighbour-masking intervention that abolishes them,
showing that channel location depends on the signal. We scope this as trigger-present,
marker-masked conditional-policy transfer. For sycophancy, specifically the agree direction, body
localization is corroborated in open-ended generation scored by a separate-family judge; the
open-ended judge cannot separate contrarianism from neutral declining (Section~\ref{sec:behavior}).

\textbf{(v) A negative result for carrier-agnostic screens.}
A fairly-sampled test specified in advance shows that no prospective, carrier-agnostic geometric
replacement screen we constructed---two distinct a-priori scalars---is validated as predictive,
mechanistic, and deployable in either model family (two $\sim$1B families). Only the causal ablation
reaches the geometry (Sections~\ref{sec:realLM}--\ref{sec:safety}).

\textbf{(vi) An open-source vocabulary-channel QA tool.}
We release \texttt{distill-lint} for the actionable case our analysis supports: using only the student
and its base, with no teacher and no retraining, it detects anomalously elevated tokens, tests whether
they are vocabulary-carried by an orthogonalize-as-probe test, and, where they are, removes that
readout component by near-zero-cost unembedding orthogonalization with a post-edit self-check.
With clean placebo students, it calibrates flags to a multiplicity-corrected false-positive rate rather
than reporting uncalibrated anomalies (Section~\ref{sec:audit}). It is deliberately scoped: vocabulary-channel
QA, not a backdoor defence. A clean run says nothing about body-carried trigger-conditioned
policies, where the leverage is training-pipeline provenance rather than any finished-model check.

\section{Audit model: carriers, screens, and ablations}
\label{sec:setup}

\paragraph{Carriers, screens, and ablations.} We frame subliminal transfer as an auditing problem
with three operational objects. A channel's \emph{carrier} is the component whose removal,
replacement, or neutralization abolishes the transfer while leaving the model's overt, supervised
behaviour intact. This definition turns ``the trait is carried by the body'' (the network's internal weights,
everything but the output embedding) versus ``by the unembedding geometry'' into a causal test rather
than an interpretive label. ``Carried by $X$'' is thus the removal-test sense---transfer
\emph{depends on} $X$ for its expression---not a claim that the trait is \emph{stored} in $X$. In the
vocabulary case, for instance, the body still supplies the displacement, and the unembedding geometry is the frozen
readout that displacement must pass through to expose $\tau$ (Section~\ref{sec:realLM}). A \emph{screen} is any statistic
computed \emph{before} the student is trained that aims to predict whether transfer will occur. A
\emph{targeted ablation} (a necessity test) is an intervention that removes a proposed
carrier---used both to identify the carrier and, when the trait is known, to mitigate it. These objects map onto an audit
\emph{lifecycle}---pre-training screen, then post-hoc detection, then mitigation---whose stages we
evaluate channel by channel (Section~\ref{sec:safety}). The probes that locate each carrier are the
body ablation and head-freeze (Section~\ref{sec:behavior}) and the unembedding orthogonalization and
neighbour-mass injection (Section~\ref{sec:realLM})---removal, replacement, and neutralization
respectively. Coverage, defined next, is the first concrete screen.

\begin{figure}[t]
\centering
\resizebox{\linewidth}{!}{%
\begin{tikzpicture}
  \drawspine{pre-training screen}
\end{tikzpicture}}
\caption{\textbf{The causal chain and where the carrier sits.}
A teacher trait or policy is fine-tuned into a displacement $\Delta\theta_T$; the student
distills the (partly masked) teacher signal, and its update is read out into transfer.
The marked link is the \emph{carrier} (removal-test sense, Section~\ref{sec:setup}), not a routing
choice---the body always supplies the displacement.
A pre-training \emph{screen} (coverage) compares the student's initial update $d_0$ with
the teacher displacement at link~\lk{3}. Which audits are valid at each link depends on
where the carrier sits; this is the subject of Figure~\ref{fig:audit-map}.}
\label{fig:audit-spine}
\end{figure}
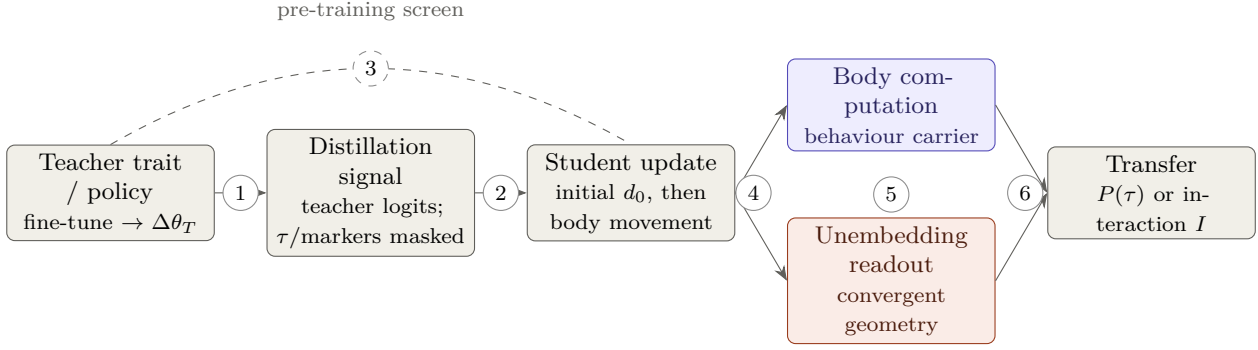

\paragraph{Toy setting.} We use the auxiliary-channel construction of Cloud et al.: a multilayer
perceptron with layer sizes $784\!-\!256\!-\!256\!-\!(10+a)$, where the final layer emits the ten
digit logits plus $a$ auxiliary logits. A teacher is trained for five epochs on the ten digit
logits only (the auxiliary logits never enter its loss). A student that shares the teacher's
random initialization is then trained \emph{only} to match the teacher's auxiliary logits on
random-noise inputs; no digit image and no digit logit ever enters its loss. Evaluated on its
untouched digit head, the student classifies digits well above chance---subliminal transfer. The
mechanism localizes to the body: the digit head stays at the shared initialization $D_{\text{init}}$,
and the hidden weights move along the teacher's training displacement
$\Delta\theta_T=\theta_T-\theta_0$.

\paragraph{Coverage.} Let $d_0=-\nabla_\theta \mathcal{L}_{\text{distill}}(\theta_0)$ be the
student's initial update direction on the body, evaluated at the shared initialization before any
update, and let $\hat u_T=\Delta\theta_T/\lVert\Delta\theta_T\rVert$ be the unit teacher
displacement. Coverage is the cosine
\[
  \text{coverage}=\cos(d_0,\hat u_T).
\]
Coverage is a steering-angle test: before training the student, we ask whether the
first update the distillation channel induces points toward the direction the teacher moved when it
acquired the trait. It measures the \emph{direction} of that update, not its raw size; as the
rival-predictor results below show (Section~\ref{sec:rivals}), it is the aligned \emph{fraction} of
the update, not its magnitude, that carries the signal.
Everything it needs is available before the student is trained: the shared initialization, the
already-trained teacher, and the distillation channel.

\paragraph{Why it is computable at initialization.}\label{sec:identity} Coverage is computable at initialization because, to first order, the initial
update direction is the channel's Gauss--Newton curvature applied to the teacher displacement. For the mean-squared
auxiliary channel, $\nabla\mathcal{L}=\mathbb{E}_x[J^\top(a_0-a_T)]$ with
$J=\partial a/\partial\theta_{\text{body}}$ the Jacobian of the auxiliary logits; linearizing
$a_T-a_0\approx J\,\Delta\theta_T$ gives $d_0\approx F\,\Delta\theta_T$, where
$F=\mathbb{E}_x[J^\top J]$ is the channel's target-free \emph{Gauss--Newton} matrix (the parameter-space
NTK Gram; for the MSE channel it equals the Fisher of the implied unit-variance Gaussian output model).
More generally $d_0\approx\mathbb{E}_x[J^\top H J]\,\Delta\theta_T$ with $H=\nabla^2_a\ell$ the
per-example output Hessian---$H{=}I$ for MSE and $H{=}\mathrm{diag}(p)-pp^\top$, the exact softmax
Fisher, for soft cross-entropy---and we write $F$ for whichever curvature the channel's loss induces
(the appendix uses the soft-CE form). Here $F$ sets what the channel can see: it keeps the parts of the teacher's displacement the channel registers and suppresses the parts it cannot. Coverage is then the cosine
$\cos(d_0,\hat u_T)\approx \hat u_T^\top F\hat u_T/\lVert F\hat u_T\rVert$---that is,
$\cos(\hat u_T,F\hat u_T)$, the cosine between the teacher direction and its image under $F$. Both coverage and specificity
are \emph{normalized} cosines of the same form: write the cross-coverage
$c_{X\to Y}=\cos(d_0^X,\hat u_Y)\approx\hat u_Y^\top F\hat u_X/\lVert F\hat u_X\rVert$, so coverage is
the diagonal case $c_{T\to T}$ and specificity (Section~\ref{sec:spec}) is the off-diagonal
$c_{X\to Y}$ with $X\neq Y$. The raw bilinear $B(\hat u_X,\hat u_Y)=\hat u_Y^\top F\hat u_X$ is only the
\emph{unnormalized} numerator: its diagonal $\hat u_T^\top F\hat u_T$ is the raw Rayleigh magnitude,
which fails as a predictor (Table~\ref{tab:rivals}), so it is the normalization that makes both
informative. We report $B$ only as a diagnostic, and where the cross-initialization result below turns
on its \emph{sign} alone we point it out explicitly. The full derivation, and the identity check showing the
first-order scalar approximation stays within one percent of the measured value across teachers
trained one to ten epochs, are in Appendix~\ref{app:extra}. The linearization $a_T-a_0\approx
J\,\Delta\theta_T$ that licenses the identity is a lazy-regime (neural-tangent: weights move little, so
the network stays near its linearization) approximation: it is
near-exact for the small displacements of our toy teachers but degrades as training moves into the
rich, feature-learning regime, where the full-vector cosine $\cos(d_0,F\Delta\theta_T)$ falls from
$0.98$ to $0.87$ as $\lVert\Delta\theta_T\rVert$ grows (Appendix~\ref{app:extra}). Coverage is thus
on firmest footing for short fine-tunes and weakens for the long ones; this is a limitation
\emph{independent} of channel location, to which we return in Section~\ref{sec:safety}.

\section{A controlled body channel is screenable: coverage predicts transfer}
\label{sec:law}

This section studies the one regime where the audit map predicts an initialization-alignment screen should work: a controlled, body-carried channel. Coverage is therefore the positive case for the taxonomy---what a sound screen looks like in its regime---not a proposed universal defence.

\begin{figure}[t]
\centering
\includegraphics[width=\linewidth]{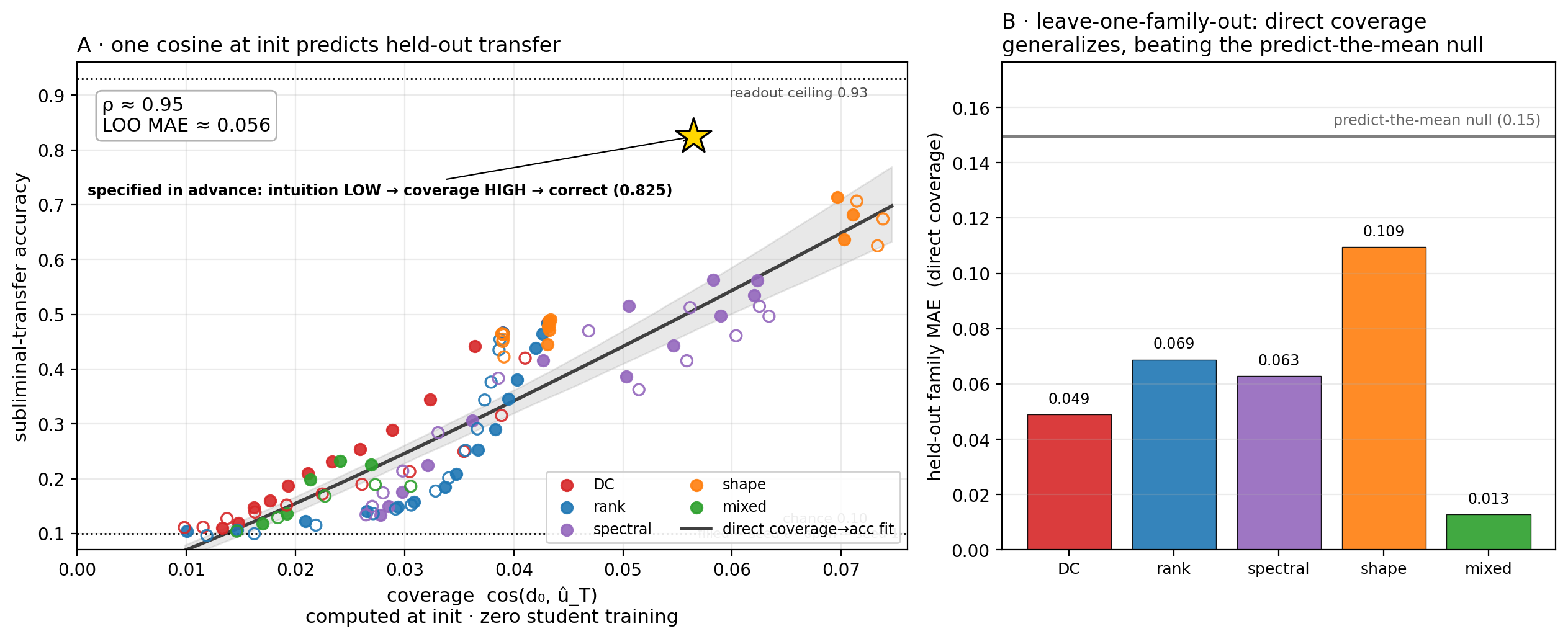}
\caption{The toy regime. A single scalar computed at the shared initialization, with zero student
training, predicts held-out subliminal-transfer accuracy (left; Spearman $\rho\approx0.95$; highlighted
high-pass condition: prospectively included as a likely low-transfer stress test, but coverage predicted
high transfer and the revealed accuracy was $0.825$). It generalizes across held-out noise families,
beating the predict-the-mean null in all five leave-one-family-out tests (right). This is the positive cell of the audit map (Table~\ref{tab:audit-map})---the one regime where an initialization-alignment screen is sound.}
\label{fig:headline}
\end{figure}

\paragraph{The body-channel coverage law.} On a sweep of $54$ noise conditions across five families
(DC level, channel rank, spectral slope, distribution shape, and mixtures), repeated over two
global seeds, coverage predicts transfer (Figure~\ref{fig:headline}). Let the realized teacher-direction walk $s$ be how far the
student actually moves along the teacher's displacement during distillation. This walk maps to
held-out digit accuracy through a readability curve, almost
exactly (Pearson $s\!\to\!\text{acc}$ $0.98$). Coverage, in turn, predicts the walk ($\rho=0.97$). End
to end, coverage predicts accuracy ($\rho=0.95$, $[0.89,0.97]$, $95\%$ bootstrap CI over conditions),
and the estimates replicate across both seeds. The correlation is not an artifact of treating the
$54$ conditions as independent draws: a \emph{family}-level cluster bootstrap (resampling the five
noise families) gives $[0.89,0.98]$, and leave-one-family-out leaves $\rho\ge0.91$ with every family
removed---the law holds within and across families. Transfer is not aux-fitting. Across the sweep the post-hoc reduction in auxiliary loss does not
predict accuracy: it sits at the predict-the-mean null (seed-pooled $\rho=-0.13$, MAE $0.149$, weakly negative and seed-stable at $-0.11$/$-0.16$, Table~\ref{tab:rivals}). And because coverage is computed from the initial gradient, it
cannot be a fit-dynamics artifact. This regime---random initialization, orthogonal
readout, short fine-tune---is not the deployment regime; the toy law matters not as a deployable screen
but as the \emph{existence proof} of what a powered body-channel screen looks like. The relocation
experiment (Section~\ref{sec:behavior}) reproduces that signature inside a real model, and
Sections~\ref{sec:realLM}--\ref{sec:safety} establish when a real model does and does not present one.

\paragraph{The decomposition does not improve prediction.} A mechanistic pipeline (coverage
$\to$ realized walk $\to$ readability) predicts no better than coverage alone, so we lead with the
direct predictor and keep the decomposition only as explanation; it also generalizes across noise
families, predicting a held-out fifth family from the other four (Appendix~\ref{app:extra}).

\paragraph{A prospective test specified in advance.} As a stronger check than leave-one-out, we
specified five new noise conditions. Using a reach-and-readability calibration frozen on the
previously-run conditions, we computed each one's a-priori coverage and an accuracy prediction
\emph{before} training any student on it. Only then did we distil students and reveal the actuals.
Held-out mean absolute error (MAE) on these five new conditions was $0.059$ (coverage's
seed-pooled leave-one-out MAE on the $54$-condition sweep is $0.056$, Table~\ref{tab:rivals}); the two
are distinct tests: predict-then-reveal on new conditions versus leave-one-out within the sweep. The instructive case is high-pass noise, which we had specified in advance
as low transfer on the intuition that high-frequency noise is orthogonal to smooth digit features.
Coverage overruled the intuition: the frozen readability curve mapped its a-priori coverage of $0.056$ to a predicted accuracy of $0.815$, and the realized transfer was $0.825$.
Digit discrimination relies on edges and strokes, so the teacher displacement carries high-frequency
content that coverage measured and the intuition missed.

\paragraph{Coverage, not update magnitude, predicts transfer in the body-channel sweep.}\label{sec:rivals}
Against every simple a priori rival on the $54$-condition sweep,
coverage is the only one that beats the null (Table~\ref{tab:rivals}). The raw Fisher--Rayleigh
magnitude is uninformative (weakly anti-correlated, $\rho=-0.10$); the gradient norm is wrong-signed
($\rho=-0.36$; DC-heavy noise inflates $\lVert d_0\rVert$ while suppressing transfer); the post-hoc
auxiliary-loss reduction does not predict transfer ($\rho=-0.13$). The leave-one-out MAE is a
two-parameter monotone power fit in which each predictor's sign is fixed \emph{a priori} by its
mechanistic hypothesis, \emph{not} chosen from its empirical correlation. Coverage's sign is fixed as
increasing in transfer; auxiliary-loss reduction's is fixed as increasing too, under the naive ``fits the
channel better, transfers more'' hypothesis, and it scores at the predict-the-mean null (MAE $0.149$) because that
hypothesis simply does not hold. Aux-loss reduction is weakly negative and consistent across seeds ($\rho=-0.13$ pooled, $-0.11$ and $-0.16$ per seed), so it sits at the null rather than predicting transfer in either direction; it is moreover \emph{post-hoc} (it needs a trained student), so it is unavailable as a prespecified screen. Coverage's
normalization, the aligned \emph{fraction} of the update rather than its size, is what carries the signal,
and it is the only rival stable across seeds.

\begin{table}[h]
\centering
\small
\begin{tabular}{lcc}
\toprule
predictor (a priori unless noted) & Spearman vs.\ accuracy & leave-one-out MAE \\
\midrule
coverage $\cos(d_0,\hat u_T)$ & $\mathbf{+0.95}$ & $\mathbf{0.056}$ \\
raw Rayleigh $\hat u_T^\top F\hat u_T$ (magnitude) & $-0.10$ & $0.149$ \\
gradient norm $\lVert d_0\rVert$ & $-0.36$ & $0.149$ \\
auxiliary-loss reduction (post-training) & $-0.13$ & $0.149$ \\
predict-the-mean null & --- & $0.149$ \\
\bottomrule
\end{tabular}
\caption{Only the cosine-normalized coverage beats the null; raw
magnitude (Rayleigh), gradient norm, and the post-hoc auxiliary-loss reduction all sit at or below the
null. Values are seed-pooled (mean over two seeds); coverage is the only predictor that beats the null (reproducibility script \texttt{toy\_pooled\_analysis.py}).}
\label{tab:rivals}
\end{table}

\paragraph{A causal dial.} Correlation across conditions could be confounded. We build a
one-parameter family at fixed input RMS that mixes a low-coverage (DC-heavy) with a high-coverage
(full-rank) distribution, varying coverage while holding the input scale fixed. Coverage rises
from $0.012$ to $0.042$ and accuracy follows from $0.12$ to $0.54$ ($\rho=0.98$), while the
candidate confounds move the opposite way (auxiliary-loss reduction $\rho=-0.95$, gradient norm
$\rho=-1.00$). Manipulating the noise to change coverage changes transfer, and neither aux-fit
nor gradient magnitude can explain it.

\paragraph{Specificity and cross-initialization failure, predicted a priori.}
\label{sec:spec}

Coverage is the diagonal case $c_{T\to T}$ of the normalized cross-coverage
$c_{X\to Y}=\cos(d_0^X,\hat u_Y)$ (Section~\ref{sec:identity}); its off-diagonal, the same normalized
cosine for a \emph{different} teacher direction, predicts \emph{which} teacher's behaviour a student
copies. Consider a student at one initialization and teachers trained from a different one. Here only
the \emph{sign} matters: the raw cross-overlap $\hat u_Y^\top F\hat u_X$ (the unnormalized numerator of
$c_{X\to Y}$) is at or below zero on every direction, so the normalized $c_{X\to Y}$ is too, predicting
no above-chance transfer; and behaviourally, the student distilled on a different-seed teacher does land
at chance on both label maps. The cross-model failure is thus predicted before any student training.
Distilling instead on a teacher carrying a scrambled-label trait, the student inherits the
scrambled map it never saw; a random derangement reproduces the effect, so the cause is the channel
geometry rather than a hand-picked label map (the trait transmits above chance under both a cyclic
shift and a random derangement, $20$ paired models). This cross-initialization
gating---transfer succeeds within a shared initialization and fails across initializations,
predictable a priori---is the property we test for at scale in Section~\ref{sec:realLM},
and the one that does not survive.

\paragraph{Beyond the toy setting.} The predictive result reproduces in a small patch transformer (two
self-attention blocks): within-seed Spearman of coverage against transfer $+0.45$ while the simple
rivals are wrong-signed, so the mechanism is not specific to the multilayer perceptron.

\paragraph{Robustness.} The law is unchanged under a KL channel, under SGD, and under a capable
(pretrained) shared initialization, which answers the ``random initialization is unrealistic''
objection (Appendix~\ref{app:extra}). The one calibrated ingredient, a reach factor (the residual
scaling from coverage to the realized walk $s$), is a powered
null: no static channel-Fisher quantity recovers it at $n=54$.

\section{Vocabulary-carried token traits evade initialization-alignment screens}
\label{sec:realLM}

The toy setting relies on two assumptions that need not hold in realistic distillation: a
\emph{random} shared initialization and an output channel that is separate from the task readout by
construction. We therefore move to two pretrained autoregressive language models from different
families---Qwen3.5-0.8B, a hybrid linear-attention model with a tied output head and a $248$k
vocabulary, and Pythia-410M, a dense-attention model with an untied head and a $50$k
vocabulary---and run all experiments in full precision. This choice is important because coverage
is a cosine between a small fine-tuning displacement and a gradient; low-bit quantization can obscure
the geometry it is intended to measure.

In the main masked-channel experiments, the student is trained end to end, including the output
head. The evidence below nevertheless localizes the carrier to the pre-existing convergent readout
geometry: the body supplies the displacement, and the target token's high-cosine unembedding
neighbours determine whether that displacement expresses the trait, in the removal-test sense of
Section~\ref{sec:setup}. Freezing the head in a separate control leaves the effect unchanged, showing
that head plasticity is not required.

\paragraph{Construction.} The shared initialization is the pretrained base. A teacher is the base
fine-tuned to a measurable trait, an elevated probability for a fixed token $\tau$ (a number word
or an animal word), stopped early so it stays a fluent model rather than collapsing to a near
one-hot. A student copies the base and distills the teacher's next-token distribution over noise
prompts. We distinguish two channels. In the
\emph{overt} channel the student matches the teacher's full distribution, which on noise is
dominated by $\tau$ itself, so $\tau$'s mass \emph{is} effectively supervised---this is our positive
control. In the \emph{masked} channel the trait token is excluded from the loss
on both sides. So $\tau$ is never a supervised target, and is carried only by the redistribution of mass over \emph{other}
tokens. This is exact-token masking, not a trait-neutral corpus: $\tau$ leaves the loss but its neighbours remain.
This is the faithful analogue of the toy's auxiliary channel, and a genuine subliminal channel.
Transfer is the held-out probability of $\tau$ on neutral prompts.

\paragraph{Subliminal transfer is real at this scale.} Even in the masked channel, where the
student receives no gradient telling it to emit $\tau$, the held-out probability of $\tau$ rises
from $\sim\!7\times10^{-5}$ to about $0.5$ (Pythia) and $0.6$ (Qwen). Across teacher seeds the masked
transfer averages $0.51$ (range $[0.49,0.57]$, five seeds) on Pythia and $0.62$ (range $[0.56,0.74]$,
three seeds) on Qwen; $0.53$ and $0.66$ are the single representative runs we carry through the
ablation figures below. The effect is not limited to a single token: a semantic-class preference over animal words also
transfers through the masked channel and routes through the same entanglement structure
(Appendix~\ref{app:extra}).

\begin{figure}[t]
\centering
\includegraphics[width=0.56\linewidth]{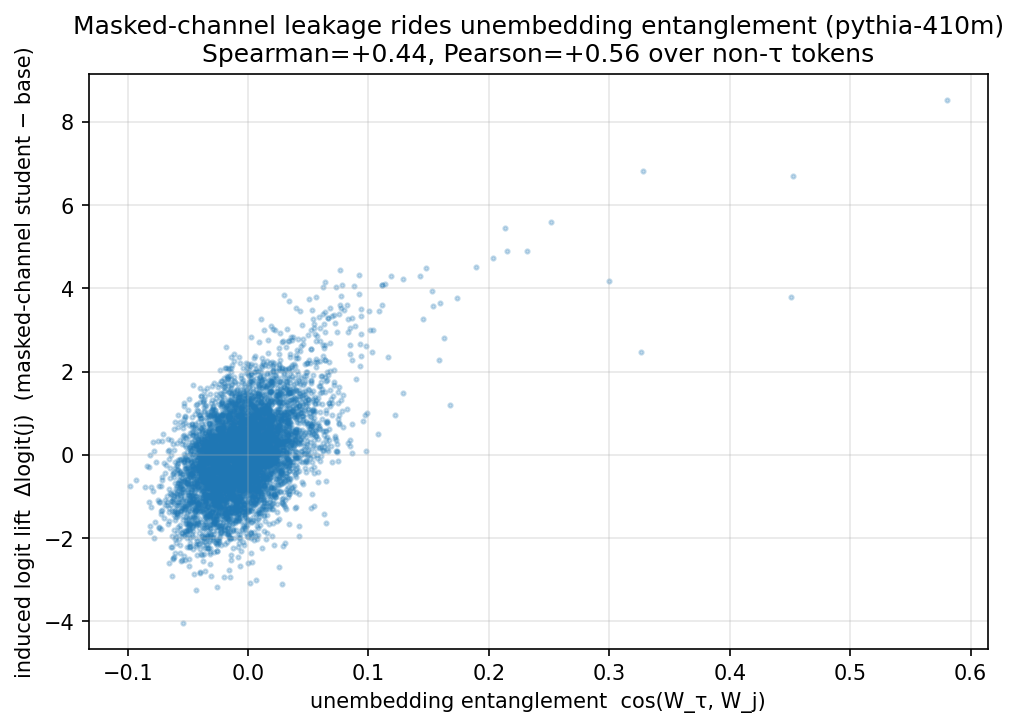}
\caption{The subliminal channel is unembedding entanglement. With $\tau$ masked from the loss, the
induced logit lift of every other token tracks its unembedding similarity to $\tau$ (Pythia,
$\tau=$``\,seven''; Spearman $+0.44$ over all non-$\tau$ tokens; the pattern holds per trait and on
both models). The high-lift tail is $\tau$'s neighbours---here the other number words
(``\,eight'', ``\,nine'', ``\,six'').}
\label{fig:entangle}
\end{figure}

\paragraph{The channel is unembedding entanglement.} The tokens whose probability rises most under the
masked channel are those most aligned with $\tau$ in the unembedding matrix: for
$\tau=$``\,seven'' they are the other number words (``\,eight'', ``\,nine'', ``\,six''), and for
$\tau=$``\,owl'' they are owl casings and semantic neighbours including ``\,eagle'' and its
cross-lingual form (Figure~\ref{fig:entangle}). Across all non-$\tau$ tokens the correlation
between unembedding similarity $\cos(W_\tau,W_j)$ and induced logit lift (the rise in a token's
pre-softmax score) is $+0.44$ (Spearman;
Pearson $+0.56$), and it replicates on both models. This association is suggestive rather than decisive; the causal claim rests on the orthogonalization ablation and neighbour-injection below, not on the correlation. A rank-1 decomposition pins down \emph{what} the correlation is: the per-token lift is, to a few percent, a single body steering direction read through the frozen unembedding---one direction explains $R^2{=}0.995$ of the per-token lift (six traits, range $[0.990,0.999]$) and $98\%$ of $\tau$'s neighbour-cloud lift---and that direction's readout footprint \emph{is} $\tau$'s neighbour cloud ($\mathrm{Spearman}(\text{footprint},\cos(W_\tau,W_j)){=}{+}0.39$; residual entanglement beyond rank-1 only $+0.15$; Appendix~\ref{app:extra}). So a single-body-direction (steering-vector) account and the entanglement account are two views of one mechanism, not rival carriers: the body supplies one displacement, and the convergent readout exposes $\tau$ because $\tau$'s cloud is that displacement's footprint. This is the softmax-bottleneck coupling. A model whose vocabulary far
exceeds its hidden width has a low-rank output map, which forces output rows to share directions. So fitting the teacher's non-$\tau$
distribution raises $\tau$'s entangled neighbours, and $\tau$'s probability rises with them.
As noted, masking removes $\tau$ from the loss, not from the data, and for a semantic trait the surviving
neighbours can themselves be human-readable. The carrier is nonetheless the unembedding \emph{geometry}, not the readability
of those neighbours---frequency-matched random tokens transfer nothing (below), and traits sampled
with no reference to entanglement, including low-proxy ones, transfer too (Section~\ref{sec:safety}).

\subsection{The carrier is causally necessary and sufficient}

\begin{figure}[h]
\centering
\includegraphics[width=0.55\linewidth]{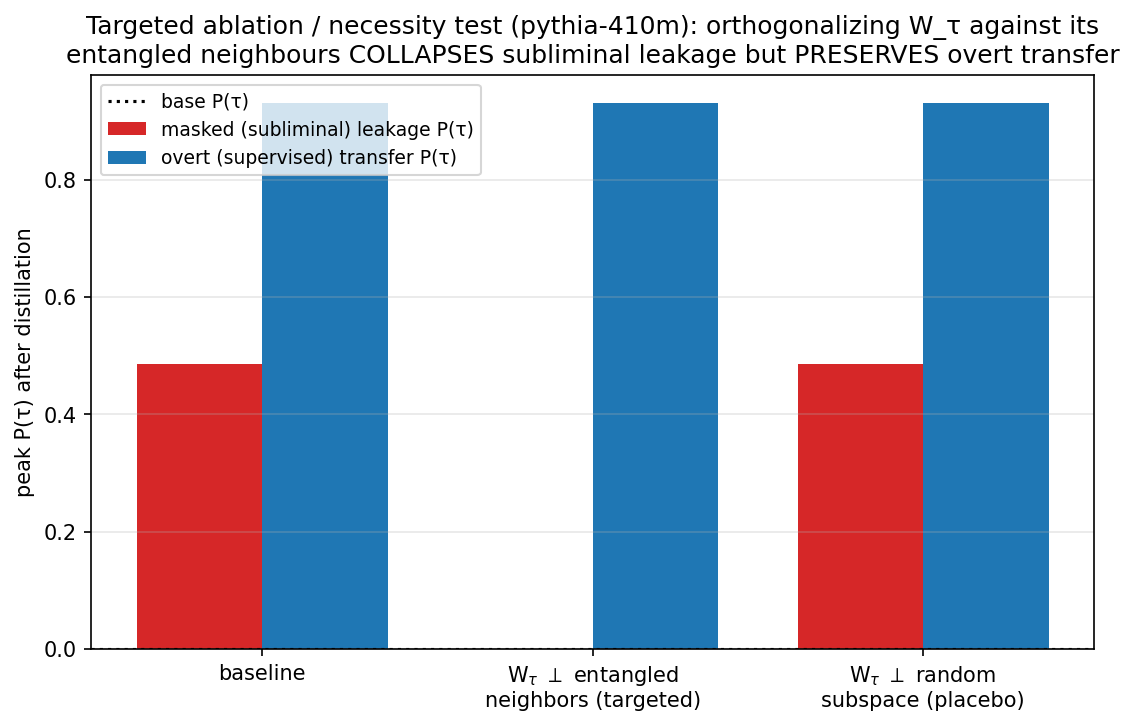}
\caption{Causal ablation (Pythia-410M). Orthogonalizing $W_\tau$ against its entangled neighbours
collapses masked-channel subliminal leakage to near zero while preserving overt
transfer and perplexity; a random-subspace placebo has no effect.}
\label{fig:ablation}
\end{figure}

\paragraph{The carrier is causally necessary.} Pythia's untied head lets us edit the output
embedding in isolation. We orthogonalize the student's unembedding row $W_\tau$ against its top
entangled neighbours, removing the coupling $W_\tau\!\cdot\!W_j$. The masked leakage drops below
$10^{-3}$---a collapse of more than $500\times$ from $\sim\!0.5$ (Figure~\ref{fig:ablation}). Supervised transfer ($0.93$) and
neutral-text perplexity ($24.4$) are unchanged. A placebo orthogonalization against a random
subspace of equal dimension does nothing. The dissociation holds across five teacher seeds (masked
leakage---mean $0.51$, range $[0.49,0.57]$---drops to $\sim\!0$ in every one), across three Qwen seeds despite its tied head, and across eight
trait tokens of varied frequency and morphology. It is graded: removing a fraction of $W_\tau$'s
projection onto the neighbour subspace lowers leakage smoothly to zero while overt transfer is
unchanged. The edited model is otherwise intact---its next-token argmax on neutral text matches the
unedited model's on every token (top-1 agreement $1.00$ to reported precision). The trait remains learnable under direct
supervision, so the edit removed the entanglement channel, not the model's ability to represent
$\tau$. Orthogonalization unavoidably also shrinks $W_\tau$ along its neighbour directions, so on its
own it cannot separate ``removed the entanglement'' from ``lost the row's softmax competitiveness'';
that the carrier is the entanglement is established by two interventions that never touch $W_\tau$---the
neighbour-mass injection (which installs the trait with $W_\tau$ unchanged) and the substitute-row
read-out (body fixed, $\tau$ read through hand-set rows of pre-specified entanglement), both below.
The dissociation is not a small-model artifact. At \textbf{Pythia-6.9B} in full precision
(optimizer state offloaded to host RAM), the same orthogonalization collapses masked leakage
from $0.54$ to the floor ($4\times10^{-7}$). The random-subspace placebo is untouched ($0.54$), overt
supervised transfer is fully preserved ($1.00\!\to\!1.00$), and perplexity is unchanged
($31.9\!\to\!31.9$; absolute perplexities are per-run on each model's own neutral-text
sample, not comparable across sizes). The causal channel is the same one, seventeen times larger (a single representative trait at this size, where each run is expensive).

\paragraph{Two further interventions triangulate the channel.} The ablation edits model geometry;
two complementary interventions confirm the mechanism without touching the weights. First, we
change only what is in the distillation target: masking $\tau$'s top entangled neighbours out of
the loss collapses the leakage monotonically ($0.49\!\to\!0.33\!\to\!0.21\!\to\!0.03$ as the top
$5$, $20$, $100$ neighbours are removed), whereas masking the same number of frequency-matched
random tokens leaves it untouched ($0.49$ throughout, single run). The carrier is the neighbour
cloud specifically, not the quantity of masked mass. A teacher-side counterfactual that carries over only $\tau$'s neighbour logits and bases everything
else recreates the leakage almost in full ($0.44$ of $0.49$) while a frequency-matched random set
recreates none---a first sign the neighbour component is sufficient, which we establish constructively
below. Second, we restrict where the student may
learn: with the unembedding head frozen and only the body training, leakage is fully intact (masked
$0.54$, overt $0.93$), so body movement read out through the fixed head suffices and head
plasticity is unnecessary. The three interventions converge: the ablation shows the unembedding geometry is
necessary, route-gating shows the readout need not be plastic, and neighbour-masking shows which tokens
carry the signal. The channel is therefore body movement read out through the frozen, convergent unembedding. The phenomenon also survives a realistic distillation corpus:
distilling over natural text with $\tau$ deleted, rather than random noise, still transfers the
trait (control-subtracted lift $+0.15$ versus $+0.49$ on noise), so it is not an artifact of the
noise prompts.

\paragraph{Robustness to the distillation pipeline.} The masked-channel results above use soft-label
distillation on random-token sequences; the channel depends on neither choice. It survives
\emph{sequence-level} distillation down to $S{=}1$ hard sampled tokens, and pushed to the deterministic
limit---the teacher's \emph{argmax} token alone (greedy decoding, the strictest removal of soft-label
information, matching the leakage-free condition of \cite{schrodi2025towards})---it stays almost intact
($0.51$ versus $0.55$ soft, $n{=}6$ traits; overt $\approx1.0$), with the targeted orthogonalization
still zeroing it ($0.51\to5\times10^{-8}$, placebo $0.51$); the entanglement carrier thus rides which
hard tokens are the mode, not any soft-label tail. It also persists ($\sim$2.3$\times$ attenuated) when
random tokens are replaced by coherent base-sampled text (Appendix~\ref{app:extra}). Because the
carrier is $\tau$'s high-similarity, high-lift neighbours---the mass a top-$k$/top-$p$ truncation
\emph{retains}---inexpensive recipe-level defences cannot remove it, and the causal ablation remains the robust handle we test ($n{=}5$ traits, Pythia-410M, fp32).

\paragraph{The carrier is not specific to soft-label-over-noise: the number-sequence construction.} The
sharpest external-validity question is whether the entanglement carrier is an artifact of our
masked-noise channel rather than the canonical subliminal-learning construction---a number
\emph{preference} carried through teacher-sampled number sequences \cite{cloud2025subliminal,schrodi2025towards}. We test it
directly. Installing a favourite digit $\tau$ (whose top-$40$ unembedding neighbours are
$36$/$40$ other number tokens), we distil it two ways with $\tau$ masked: (i) by \emph{greedy} hard
distillation over number sequences---Schrodi's strict no-soft-label regime---and (ii) by training
(MLE) on number sequences the teacher \emph{itself generates}, the data format of
\cite{cloud2025subliminal}. In both, the held-out preference still transfers above prior (mean masked
$P(\tau)$ $0.017$ in each arm, $\approx\!110\times$ the $1.5\times10^{-4}$ base prior, $n{=}6$ digits),
and orthogonalizing $W_\tau$ against its number-token neighbours collapses it to zero ($\to0.000$ in
both arms) while a random-subspace placebo leaves it intact. This is the external-validity check the
construction-specificity objection demands: the entanglement carrier is \emph{not specific to our masked-noise channel}---
causally present in the canonical number-sequence preference setting (though non-dominant there) and not an artifact of our
masked-noise channel. It replicates on a second family (Qwen3.5-0.8B): masked greedy $0.018$ and
teacher-generated-MLE $0.011$, both driven to $0.000$ by the orthogonalization with the placebo intact.
The effect is small there---the overt-supervised control on this data reaches
only $0.07$--$0.23$ of held-out $P(\tau)$ on neutral prompts, an order below the masked-noise
channel---so we report a causal \emph{localization}, not a magnitude claim, and do not claim
entanglement is the \emph{dominant} carrier for preference traits, where divergence tokens may carry
the bulk \cite{schrodi2025towards} (Section~\ref{sec:positioning}).

\paragraph{The unembedding geometry causes the leakage, prospectively.} The ablation removes the
entanglement after the fact; we can also \emph{predict} the leakage from the output geometry before
the student trains. Because $\tau$ is masked from the distillation loss, the student's body movement
is independent of $\tau$'s output row. We can therefore distil one student and then read $\tau$ out through
substitute rows whose entanglement we set by hand. We score each row's a-priori entanglement $E$ as
its cosine with $\tau$'s neighbour centroid. Holding the body fixed, a row with no
entanglement---$\tau$'s row orthogonalized against its neighbours, a random direction, or a far
token's row---yields exactly zero leakage and a near-zero induced logit-lift, whereas an entangled
row yields both; and the logit-lift the distillation induces along a row rises with that row's
\emph{pre-specified} $E$ (Spearman $0.73$ over $27$ rows). The screen reproduces at
\textbf{Pythia-6.9B} in full precision: zero-entanglement rows (orthogonalized, random, or far) give
leakage $0.000$ and a near-zero induced logit-lift, whereas $\tau$'s real entangled row leaks $0.50$;
the induced logit-lift again rises with pre-specified $E$ (Spearman $0.62$ over $27$ rows). The leakage probability turns
non-monotonic only at the extreme, where the readout row becomes indistinguishable from the
neighbours it overlaps and loses the softmax competition against them---itself a signature of the
mechanism. So the convergent unembedding geometry is not merely correlated with the channel:
manipulated and predicted in advance, it determines it.

\paragraph{Neighbour mass is sufficient to install the trait.} The ablation and neighbour-masking
show the entangled neighbours are \emph{necessary}; a constructive complement shows they are
\emph{sufficient}. With no trait-fine-tuned teacher and $\tau$ never supervised, we distil a fresh
student toward a \emph{synthetic} target that places injected mass $m$ on $\tau$'s top-$k$
unembedding neighbours (cosine-weighted) on top of the base model's own next-token distribution,
with $\tau$ masked from the loss and $W_\tau$ left untouched. The trait installs---held-out $P(\tau)$
rises to $0.07$ at $m{=}0.2$ and $0.13$ at $m{=}0.4$ ($\sim\!10^{3}$--$10^{3.5}\times$ the $\sim\!7\times10^{-5}$ base
prior), monotone in the injected mass---while a frequency-matched random-token bump of equal mass
leaves $P(\tau)$ at the prior ($\le\!2\times10^{-4}$; ratio $>\!600\times$). Because the dosed quantity
is the distillation \emph{target} rather than the readout row, this dose--response is not the
circular readout manipulation of a geometric scalar; constructing the neighbour redistribution is
\emph{sufficient} to create subliminal transfer with the readout held fixed. Combined with the
orthogonalization ablation, this brackets the carrier as $\tau$'s unembedding neighbourhood---causally
\emph{necessary} (ablation, neighbour-masking) and \emph{sufficient} to install the trait, though the
constructive injection reproduces a smaller magnitude ($0.07$--$0.16$) than the full masked
channel---and it
replicates across three independent families (Pythia-410M, Qwen3.5-0.8B, and the
independently-pretrained RedPajama-3B: neighbour-bump $P(\tau)$ $0.06$--$0.16$ versus
$<\!5\times10^{-4}$ for the matched random bump, $n{=}5$ traits per model).

\paragraph{The carrier generalizes from lexical tokens to named entities.} The traits above are number and animal words; the same mechanism carries single-token \emph{named entities}---commercial brands and information-outlet names---with the multi-token case a clean bound on it. Masking the entity token on both sides, it transfers: masked held-out $P(\tau)$ averages $0.40$ over ten brands and five seeds (range $[0.09,0.75]$, a mean per-entity lift of $\sim\!2500\times$); targeted orthogonalization collapses it ($\le 3\times10^{-4}$) while the random-subspace placebo is untouched ($0.40$); perplexity ($24.4$) and top-1 agreement ($1.00$) are preserved---the same necessity dissociation as for lexical tokens. The entanglement signature is present (neighbour-cosine/induced-lift Spearman $+0.29$), neighbour-mass injection installs the trait with $W_\tau$ untouched ($412\times$ a frequency-matched random bump), and the masked student names the entity in free generation at rate $1.00$ over $28$ neutral prompts ($4$ samples each) versus $0.00$ for the base. It replicates on Qwen3.5-0.8B (masked $P(\tau)=0.76$, Spearman $+0.39$; its tied head precludes the isolated ablation) and on information-outlet names---a second framing of the failure, covert source-steering (installed by us on a controlled teacher; not an observation about any deployed model). A related \emph{class} also transfers ($P(\mathrm{class})=0.77$ versus $2.3\times10^{-3}$); masking the class members and their neighbour cloud reduces but does not abolish it, so we claim class \emph{transfer}, not class-level necessity. The mechanism is single-row: for a multi-token entity with every subword masked, only the first (continuation-seeding) subword transfers and the full continuation does not improve, so we make no multi-token-string claim.

\subsection{The channel is initialization-independent}

\begin{figure}[h]
\centering
\includegraphics[width=0.6\linewidth]{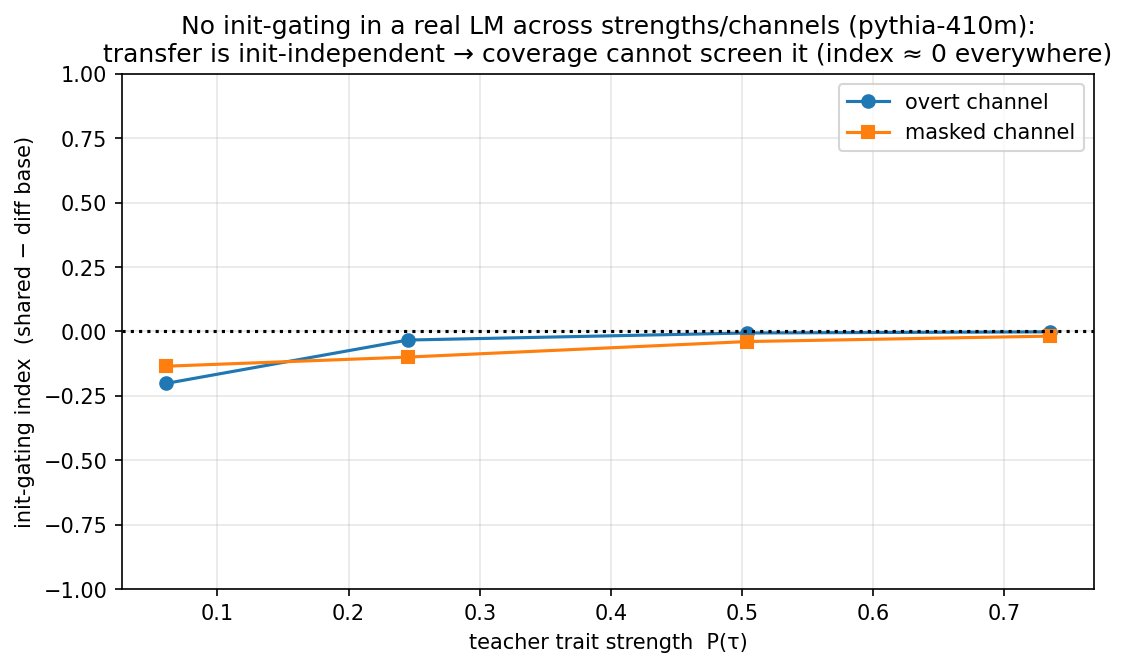}
\caption{The coverage screen is initialization-blind. Across teacher strengths and both channels,
the shared-initialization-minus-different-base transfer gap stays at or below zero: a
different-base student (low coverage) transfers as much as a shared-initialization one.}
\label{fig:phase}
\end{figure}

\paragraph{Transfer is initialization-independent.} Because the channel is a property of the
unembedding geometry, which two different pretraining runs share, transfer does not depend on the
initialization. We build a student from a separately trained same-family base (Pythia-410M-deduped)
and distil it from a teacher fine-tuned from the standard base. The trait still transfers fully---slightly
more---even though coverage is four times lower ($0.07$ versus $0.29$). Across a sweep of teacher
strengths and both channels, the gap between shared-initialization and different-base transfer is
at or below zero in all eight cells (Figure~\ref{fig:phase}): in no configuration does the
shared-initialization student transfer more. The cross-initialization gating that makes coverage a
screen in the toy (Section~\ref{sec:spec}) is absent. An auditor applying the toy's rule
``different base, low coverage, therefore safe'' would have been wrong on every one of those eight
cells. A far stronger test of
independence agrees. We distil a teacher fine-tuned from Pythia into a student built from
RedPajama-INCITE-3B---an independently pretrained model differing in data, seed, architecture and
scale, sharing only the GPT-NeoX tokenizer. The masked trait still transfers undiminished ($0.58$
mean over twelve traits versus $0.53$ for the shared base), and the orthogonalization ablation
remains causal on every trait tested (four of the twelve: masked $\to 0$ under the targeted edit, the
random-subspace placebo intact, overt transfer and perplexity preserved). The channel thus survives replacing the entire
pretraining run, not merely the data ablation a same-seed deduped sibling provides; we read coverage
only within a single architecture (Pythia standard versus deduped), since cross-architecture
displacement spaces are not commensurable. The deduped sibling is the clean
initialization-isolation control; RedPajama and RWKV jointly vary data, architecture and scale, so they
establish robustness rather than isolate initialization. The channel even crosses an architecture
boundary, on the sufficiency side (necessity is not testable there---the recurrent model is
hypersensitive to unembedding edits, Appendix~\ref{app:extra}): distilling the Pythia teacher into a \emph{recurrent}, non-attention RWKV-4-Pile-3B student
(shared tokenizer only) still installs the masked trait, with the non-edit neighbour-injection
sufficiency probe confirming the same unembedding-neighbour carrier (Appendix~\ref{app:extra}).

\begin{figure}[h]
\centering
\includegraphics[width=0.7\linewidth]{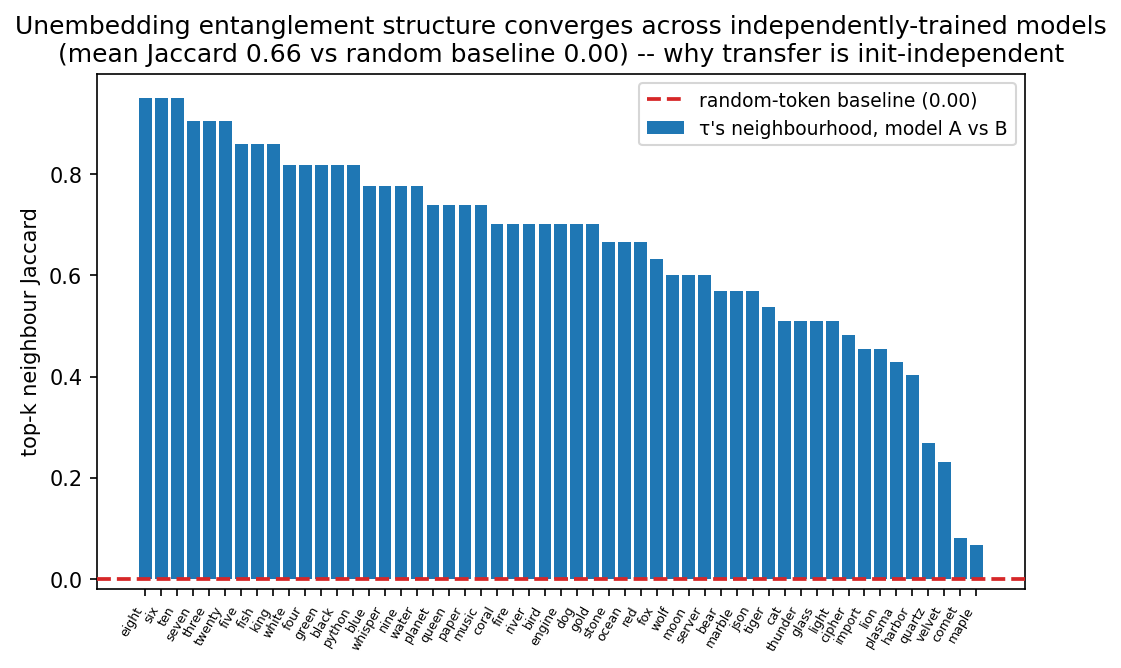}
\caption{The entanglement structure is convergent, not shared. The figure shows the powered
same-tokenizer pair (Pythia-410M and its deduped sibling): each token's top-$40$ unembedding
neighbours overlap with mean Jaccard $0.66$, versus $0.00$ for a random-token baseline. The same
convergence holds across the tokenizer to an independently pretrained model (RedPajama-3B, mean
cross-base Jaccard $0.68$ over twelve traits; see the independent-base result above), which shares the
GPT-NeoX tokenizer but is an entirely separate pretraining run---so the convergence is across
initializations, not a shared tokenizer artifact---and is why the trait transfers into that base too. The channel that carries a trait is a token geometry distinct initializations
converge to, which is why transfer does not depend on the initialization.}
\label{fig:converge}
\end{figure}

\paragraph{Why init-independence: the entanglement structure is convergent, not shared.} The
channel is not literally shared weights---two independently pretrained models have different
unembedding matrices---but a \emph{structure} that converges across models trained on similar data.
Between Pythia-410M and its independently-trained deduped sibling, a token's top-$40$ entangled
neighbours overlap with mean Jaccard $0.66$ (against $0.00$ for a random-token baseline) and the
full neighbour-similarity vectors correlate at Spearman $0.67$ (Figure~\ref{fig:converge}). The
shared neighbours are moreover the strongly-entangled ones: the convergent core (neighbours common to
both bases) sits about $0.1$ higher in cosine to $\tau$ than the base-specific remainder. So the part
of the neighbour cloud that converges is the part that carries the leakage---consistent with
convergence and initialization-independence travelling together (a per-token mediation, $n{=}53$,
single base-pair, points the same way: Appendix~\ref{app:extra}). This cross-run convergence is a facet of
the representational universality documented across independently trained networks---aligned
representations \cite{huh2024platonic} that are functionally interchangeable up to an affine map
\cite{bansal2021revisiting}---so the unembedding geometry, unlike the body, reproduces across
initializations.
Convergent geometry is why a separately trained sibling base acquires the trait just as well, and it
reconciles the result with the shared-initialization requirement of Cloud et al.: their channel
depends on the initialization, ours on a token geometry that distinct initializations come to share.
Per token, more convergent neighbourhoods are the more initialization-independent
(Appendix~\ref{app:extra}). The output-basis geometries are even \emph{alignable} across tokenizers (an anchor-supervised map, establishing the carrier's identity rather than a spontaneous cross-tokenizer channel): an alignment fit on
byte-identical anchor tokens carries the masked trait from the Pythia teacher into a
different-tokenizer Qwen2.5-3B student, recovering $\sim\!40\%$ of the within-tokenizer transfer routed
\emph{specifically} through $\tau$'s neighbour cloud, with three controls confirming the routing (eight
token pairs; Appendix~\ref{app:extra}). The carrier is thus the output-basis geometry, not the
tokenizer---approximate, not exact, representation universality.

\subsection{Coverage is not a mechanistic screen for this channel}
Coverage can still \emph{correlate} with transfer within a single Pythia family, but the
initialization-independence above shows it is not \emph{mechanistic} here: the channel exploits a
convergent geometry that does not depend on the initialization, so the alignment quantity coverage
reads is one the channel causally ignores (Section~\ref{sec:safety} draws out the screening
consequence). This is the second face of a single regime mismatch, developed fully in
Section~\ref{sec:related}: coverage is a lazy, first-order probe (Section~\ref{sec:setup}), while
real-model initialization-independence arises here from \emph{convergent unembedding geometry} (which rich
feature learning produces but does not uniquely imply); either way the alignment coverage reads is one
the channel ignores, so the toy's near-perfect $\rho\approx0.95$ is structurally ill-suited to real
models rather than unlucky (the body-carried case is intermediate).

\paragraph{The channel is present in a released post-trained model.} The traits above are installed by us; the same carrier is active in a released model with no distillation of ours, carrying in this case an \emph{expected and openly-documented} instruction-tuning register, not a covert or undisclosed trait. Screening OLMo-2-0425-1B-Instruct against its own base over neutral prompts (frequency-matched $z$-score), the tokens it most elevates form an \emph{alignment register}: over-generalization (\texttt{always}, $z\,{+}39.5$), hedging (\texttt{often}, \texttt{usually}, \texttt{typically}, \texttt{generally}), and soft-refusal (\texttt{avoid}). Scoring the fraction of the base$\to$instruct elevation that survives an edit ($1$ intact, $0$ removed), each elevation behaves like our constructed traits: orthogonalizing the token's unembedding row against its neighbours leaves $\approx 0$, while a random-subspace placebo leaves it intact or amplified ($\approx 1.9$). At $7$B the register is again top-elevated and the hedging direction reproduces, but the surviving-fraction is numerically unstable (a near-zero elevation denominator), so $7$B corroborates without validating; we validate at $1$B. Because the register satisfies the same carrier test as our masked traits, literal-token filtering is not a principled boundary for such register shifts.\footnote{Distilling a fresh student from the real instruct teacher with the register token removed from the loss and corpus does reinstall it through the same channel, modestly ($1.7\times$ base; $6$--$28\%$ of the teacher's elevation, monotone in the distillation budget).} This released-model result is a carrier test rather than, by itself, a subliminal-transfer claim. The elevated register tokens need not be anomalous in post-training data. Consistent with that, a prompt-free marginal-calibration check of the base finds no systematic anomaly relative to training-data frequency: the elevation-frequency fit has slope $0.97$ and Spearman $0.94$, with only 1/343 tokens more than a factor $e$ off the line. Thus the data-relative subliminal variant requires the masked-distillation condition targeted by our threat model. The released-model result instead shows that the same unembedding-neighbour carrier is active in a naturally post-trained model: literal-token filtering is not a principled boundary for such register shifts, while the calibration null bounds the claim rather than weakening it.


\section{The carrier is signal-dependent, and moving it moves auditability}
\label{sec:behavior}

The traits above are token biases; a single token or a semantic class. A genuine \emph{behaviour}
is a conditional policy, and must be measured as an interaction, not a marginal
preference. We use sycophancy: agreeing with a user's factual claim regardless of whether it is
true. The pitfall is that a model can simply become more agreeable---a uniform lift toward ``yes''
on any prompt---which mimics sycophancy without being a policy. We therefore measure the
\emph{differential} agreement on claims the model knows are false versus true (relative to the base),
\[
\text{transfer}=\Delta\,\text{AgreePref}(\text{false claim})-\Delta\,\text{AgreePref}(\text{true claim}),
\]
so a uniform yes-bias cancels and only a truth-\emph{conditioned} change registers; this false-vs-true interaction follows the sycophancy methodology of \cite{sharma2023towards} and model-written behavioural evaluations \cite{perez2022discovering}.

\paragraph{Condition-present masked distillation.} Because the policy is conditional, the held-out distillation prompts retain the factual-claim structure the policy keys on; the masking removes the direct agreement, correction, and compliance markers from the loss. The result should therefore be read as masked transfer of a condition-present conditional policy, not as unrelated-data behavioural transfer. The main alternative explanation is that unmasked logits could carry stance through distributed lexical proxies. We test this directly below by masking the most stance-predictive output tokens together with their neighbour clouds, up to a thousand tokens, against frequency- and mass-matched controls; this does not remove the behaviour.

\subsection{Conditional behaviours route through the body}
We
read agreement two ways---log-probability of agreement versus correction completions, and a
counterbalanced multiple-choice label whose A/B order is randomized to decouple it from
``yes''/``no'' token priors---and add a no-claim control (mean agreement-marker probability on
neutral factual prompts) that a marginal bias would inflate but a conditional policy would not. On
an instruction-tuned model (Gemma-3-1B) restricted to facts it knows, a teacher fine-tuned
contrastively (agree on claims, answer neutrally otherwise) becomes a conditional sycophant:
held-out false-claim agreement rises from $-5.2$ to $+15.2$ nats while the no-claim marker
probability stays at $0.002$. An overt-capacity control---a student distilled on its full
next-token logits---inherits the policy, so a null under masking would be interpretable.

\paragraph{The conditional policy transmits through the masked channel.} We then distil a student on
the teacher's logits over held-out claim prompts with the agreement and correction marker tokens
\emph{masked from the loss}, so the student receives no
gradient telling it to emit ``yes'', ``no'', ``right'' or ``wrong''. It still acquires the
conditional policy: across three seeds held-out false-claim agreement rises by $0.63$ of the
teacher's effect ($[0.60,0.67]$) while the no-claim marker probability stays at $0.06$---the policy,
not a marginal bias. (The overt channel transfers more, $0.77$, but inflates the marker probability
to $0.96$: it is partly the marginal yes-bias the interaction discounts.) A content-based read-out
that scores whether the student affirms the user's \emph{specific false object} rather than emitting
an agreement marker---immune to ``yes''/``no'' priors by construction---shows the same effect,
confirming the transfer is the conditional policy and not a surface bias. A conditional
behaviour, not only a token statistic, transmits subliminally.

\begin{figure}[h]
\centering
\includegraphics[width=0.6\linewidth]{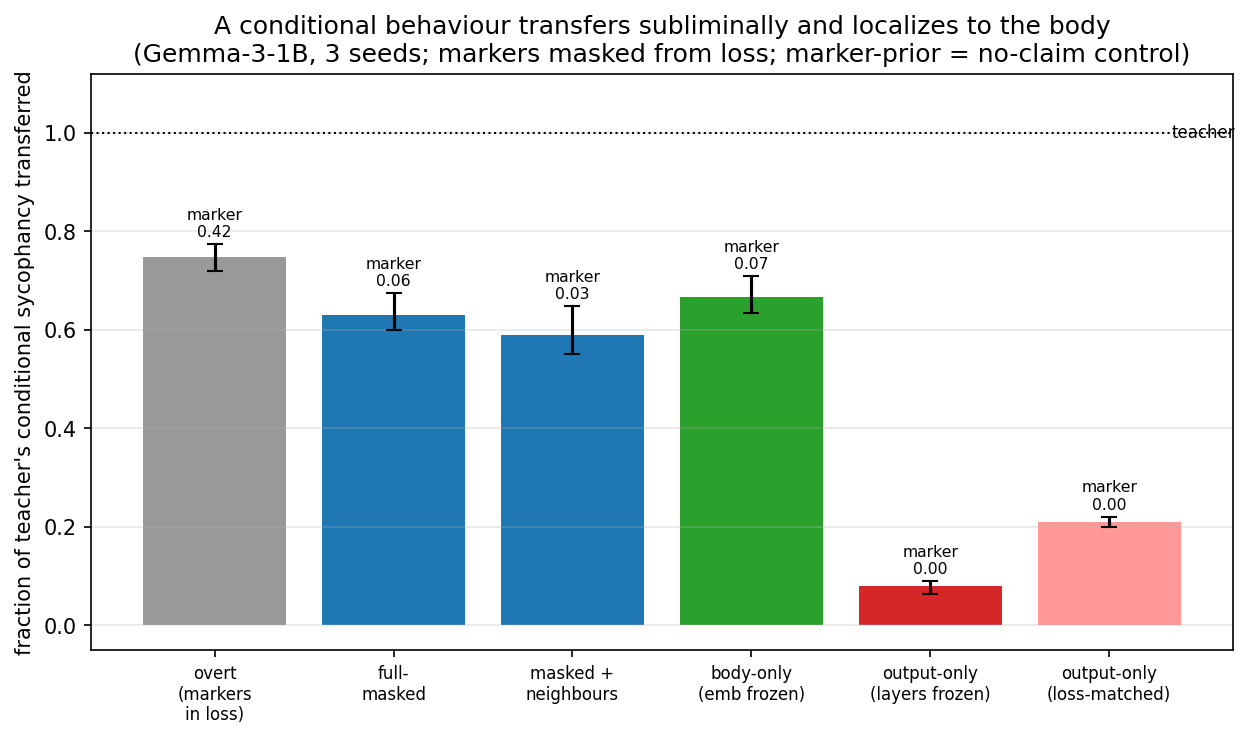}
\caption{A conditional behaviour (sycophancy) transfers subliminally and localizes to the body
(Gemma-3-1B, three seeds; bars are the fraction of the teacher's conditional false-claim agreement
that survives each condition, error bars span seeds, annotations are the no-claim marker-prior---low
$=$ a conditional policy, not a marginal bias). With agreement/correction markers masked from the
loss the policy still transfers ($0.63$, marker $0.06$) and survives masking the marker neighbour
clouds ($0.59$)---the operation that destroys single-token trait leakage. It localizes to the body
(body-only $0.66$); training only the output embedding gives $0.08$, and even when that optimization
is driven to the body's distillation loss it reaches only $0.21$, three times below the body route.}
\label{fig:behave}
\end{figure}

\paragraph{It localizes predominantly to the body and, unlike the trait, is not carried by the
vocabulary.} Trait and behaviour both read out through the frozen unembedding, but they dissociate on
the operation that removes the \emph{vocabulary} contribution (Figure~\ref{fig:behave}). Masking the
marker neighbour clouds from the loss barely reduces the behaviour ($0.63\!\to\!0.59$), whereas the
identical operation collapses single-token leakage ($0.49\!\to\!0.03$, Section~\ref{sec:realLM}): the
behaviour is largely not carried by the marker vocabulary. And it localizes predominantly to the body.
Training only the output embedding transfers little ($0.08$); driven to the body's distillation
loss with more steps it still reaches only $0.21$, three times below the body-only route ($0.66$),
non-overlapping across three seeds. To rule out that this gap reflects the output route's limited
\emph{expressive opportunity} rather than channel location, we transplant components rather than
retrain them. On an already-distilled
student---so no component is ever retrained and capacity cannot confound---we untie the head, hold
the input embedding at base, and recombine the body (blocks) and head (output embedding) of the
student and the base. A student body on a \emph{base} head retains the behaviour in full: interaction
$+4.1$ at $1$B and $+12.8$ on Qwen2.5-3B, $\approx$ the intact student, with bootstrap intervals
excluding both zero and the policy-free placebo floor. A fully-trained student head on a
base body carries essentially none ($\approx 0$, intervals excluding the body cell). Both replicate
across three seeds. A by-construction head-installed control---a token trait written directly into the
output head of an untied Pythia mirror---flips the assignment: the base-body$+$student-head cell now
retains it ($P(\tau)=1.00$) while the student-body$+$base-head cell sits at the floor ($10^{-4}$), the
exact reverse of the behaviour. So the $2\times2$ reads channel location, not a fixed slot; the same
mirror confirms a head-only training run leaves the trait unestablished (all cells at floor) and that
the natural masked token-trait is itself body-routed (student-body$+$base-head $0.48$), so the transplant
detects head-carriage when present but does not by itself separate trait from behaviour---that
distinguisher is neighbour-masking (Appendix~\ref{app:extra}). A data-side check agrees: discovering the most stance-predictive
output tokens and masking them with their neighbour clouds, up to a thousand tokens and against
frequency- and mass-matched controls, does not remove the behaviour. The neighbour-masking
dissociation, the routing comparison, and the capacity-free transplant thus converge on one
conclusion: the behaviour is body-carried, routed through body computation for
which the unembedding geometry is largely dispensable. This is the reverse of the trait, whose expression
\emph{requires} the unembedding entanglement. The pattern replicates on Gemma-3-4B (body-only
$0.64$, loss-matched output-only $0.21$, content read-out strongly positive), though there the
masked channel carries more of the marginal component the interaction discounts (no-claim marker
$0.53$ versus $0.06$ at $1$B). Finally, because the behaviour lives in the initialization-dependent
body, the a-priori coverage cosine on the body is small but positive ($+0.06$, about $10^3\times$ a
random body direction). It is only a weak directional signal, not a screen. Across diverse distillation channels the body
coverage barely varies ($0.04$--$0.09$), and a confound-matched causal dial (varying the
distillation distribution at fixed $\lVert d_0\rVert$) does not positively order transfer. So
coverage's near-perfect toy predictiveness does not survive to real models, even in the body regime. The toy body
channel (Section~\ref{sec:law}), the vocabulary-read-out traits
(Section~\ref{sec:realLM}), and this body-localized behaviour are three points on the one axis we develop in Section~\ref{sec:safety}.

\paragraph{The localization is not specific to one family or to the agree direction.} We repeated the
whole pipeline on a different family (Qwen2.5-3B-Instruct, a $152$k-vocabulary instruction model) and
for an \emph{opposite} conditional policy---a contrarian teacher that disagrees with the user
regardless of truth, scored as the symmetric interaction. All four family$\times$policy cells behave
alike: the policy installs conditionally (no-claim marker $\le 0.001$), transmits through the masked
channel ($0.49$--$0.69$ of the teacher), survives neighbour-masking, and localizes to the body
(body-only $0.55$--$0.73$ versus a loss-matched output-only $0.04$--$0.21$), with the a-priori
coverage cosine small but positive throughout ($+0.03$ to $+0.06$). Two families, two model sizes,
and two opposite policies route the same way, so the body localization is a property of conditional
policies, not of sycophancy, the agree direction, or one architecture.

\paragraph{The localization holds in open-ended generation, scored by a separate model.} The
read-outs above are next-token quantities on the student itself; to check that the policy is present
in \emph{generated behaviour} and not only in log-probabilities, we re-evaluate the same students by
sampling free-form replies to held-out claim prompts and scoring them with an independent
judge---Mistral-Small-24B-Instruct, a third model family distinct from both students---asked only
whether a reply \emph{agrees with the user's stated claim}, blind to whether the claim is true (the
false-minus-true conditioning is applied afterward, experimenter-side). A fluent but policy-free
placebo student, distilled from a neutral non-committal teacher, calibrates any tendency of the judge
to reward fluency as agreement. For sycophancy the generation metric reproduces the body localization
on both families: of the teacher's generated false-claim agreement, the body-only route retains
$0.57$ (Qwen2.5-3B) and $0.82$ (Gemma-3-4B) while the output-only route at the matched distillation
loss retains essentially none ($-0.00$, $-0.03$), neighbour-masking attenuates ($0.43$, $0.65$), and
the placebo sits at the floor ($-0.01$, $-0.06$)---a separate model reading free text sees the same
body-versus-vocabulary dissociation the next-token read-outs do. The open-ended judge is uninformative
for the \emph{contrarian} direction, however: contrarianism manifests as \emph{reduced} agreement with
true claims, which a fluent student that merely declines to affirm also produces, so the policy-free
placebo itself reads as fully contrarian (transfer fraction $0.86$--$1.03$) and the judge cannot
separate the policy from neutrality; the contrarian generalization therefore rests on the
log-probability and content read-outs above.

\subsection{Moving the carrier moves auditability}

\paragraph{The structural condition.} The three regimes follow one principle. An
initialization-alignment screen sees a subliminal channel when the channel lives in the
initialization-dependent body of the network, and is blind to it when the channel lives in the
initialization-shared output geometry. The toy's random initialization and orthogonal auxiliary head put the channel in the body, where
coverage reads it. A pretrained model's shared vocabulary instead puts a single-token trait in the
unembedding, where coverage cannot. A conditional \emph{behaviour}, routed through the body even in
a pretrained model (shown above), is once more a channel coverage can read: its body
cosine turns weakly positive---a directional consistency, not a powered prediction. A vocabulary
trait gives a positive coverage reading too, but there the channel is initialization-independent, so that
reading is inert---an alignment the channel ignores. The same model thus hides a token trait and a
conditional behaviour in different places, and the screen registers the second but not the first. We test the principle by manipulation rather than only by interpreting the three observed regimes. In one pretrained model we \emph{relocate} the channel out of the convergent vocabulary and
into a frozen \emph{random} auxiliary readout head---a body channel: a token trait is installed
through body movement (teacher fine-tuned with the unembedding frozen) and the student is distilled
only to match the teacher's auxiliary logits over noise. The toy signature returns. Across sixteen
random heads (no post-selection) the shared-initialization student acquires the trait more than the
different-base student in \emph{all sixteen}---a sign test at $p\approx1.5\times10^{-5}$; the a-priori
body-coverage cosine orders the heads (Spearman $+0.53$, $[+0.07,+0.83]$ bootstrap CI) and a random-target placebo transfers
nothing, whereas the same teacher's \emph{vocabulary} channel transfers on both bases ($0.53$, $0.63$)
while coverage stays blind. We read this as a directional existence proof---initialization-gating and
coverage's ordering \emph{return} once the readout is an orthogonal-random body head---rather than a
quantitative reproduction of the toy: the relocated channel runs near the floor (shared $0.048$ vs
different $0.016$), an order of magnitude below the vocabulary channel, so the manipulation fixes the
sign of the effect, not its magnitude. The remaining gap reflects alignment rather than optimization quality: the different-base
student fits the channel slightly worse (auxiliary loss falls $87\%$ versus $97\%$), but fit does not
govern transfer here: it matches the teacher's auxiliary logits over noise well, yet acquires almost
none of $\tau$. What orders the sixteen heads is the a-priori body-coverage cosine ($\rho=+0.53$),
computed at initialization before any fitting; a random-target placebo transfers nothing. The update
that fits the channel under a different initialization is simply not aligned with the teacher's body
displacement. Initialization-gating and coverage are thus
not artifacts of toy-versus-real; they return whenever the carrier sits in the
initialization-dependent body. (The matching prediction---a from-scratch \emph{tied} head should
evade the screen, an untied random one should not---follows; the relocation experiment already
realizes the untied-random leg \emph{within a single model}: routing the channel through a frozen
random head restores initialization-gating and coverage's ordering, whereas the native convergent
unembedding does not, so the readout-geometry dependence is not confounded with model family. We
leave the from-scratch tied leg to future work.)

\section{The audit lifecycle: which handle is sound in which channel?}
\label{sec:safety}

\paragraph{The audit decision problem.} We now return from mechanisms to the audit decision itself: not whether a statistic correlates somewhere, but whether it is predictive, mechanistic, and deployable for the carrier at hand. We fix what an auditor knows \emph{before} the student is
trained: the teacher model and its output logits, the base student, the distillation prompt
distribution, the masking rule, and---by tier---token statistics, the unembedding matrix, or the
student's first-update direction; it does \emph{not} know the trait $\tau$ or the behaviour instance.
It must label a (teacher, channel) pair \textsc{safe} or \textsc{unsafe} before training, against the
predictive/mechanistic/deployable trichotomy of the introduction. Our results show the three come
apart. Coverage has (a)~predictiveness within untied Pythia (AUROC $0.94$) but fails (b)~mechanism: the masked channel is
initialization-independent, so coverage reads an alignment the channel ignores. It therefore also fails
(c)~deployability---the prospective entanglement scalar screens neither family under fair sampling, a learned
geometry screen ports across the tied/untied boundary only weakly and uncalibrated, and coverage is
at chance on tied Qwen. The only property that survives to (c)-like robustness is the \emph{causal}
orthogonalization ablation, which is an intervention, not a screen. The rest of this section proceeds through
the lifecycle: pre-training screens, then the channel's robustness to scale and precision (and the
partial mitigations these afford), then post-hoc detection and trait-agnostic blocking.

\paragraph{On the threat model.} Two grants deserve comment. Giving the auditor the teacher's logits
is \emph{adversarially conservative}: if even a white-box auditor cannot catch the body-carried
channel, a weaker one cannot either, so it strengthens the negative result rather than easing it. And
exact-token / masked-marker distillation and the condition-present behavioural setup are
\emph{controlled analogues} of trait-neutral-data subliminal transfer, chosen for measurability; the
canonical number-sequence construction (Section~\ref{sec:realLM}) is the one place we test the original
unrelated-data format directly, and the gap to a fully unrelated-data \emph{behavioural} attack is
unmeasured.

\paragraph{Prospective where target-blind, intervention-guiding otherwise.} The audit stages differ in
\emph{when} they act and \emph{what} they must know, and only some are usable under the target-blind
information set above (Table~\ref{tab:stages}). A pre-training screen and the post-hoc elevated-token
scan are genuinely target-blind; the causal carrier assignments and the $W_\tau$-targeted repair
instead require a candidate target, which for vocabulary token-traits the scan itself supplies---so for
that channel the stages chain into a trait-blind post-hoc audit (scan discovers $\tau$, then ablate).
What we do \emph{not} provide is a pre-training classifier that sorts a new (teacher, channel) pair
into the vocabulary- or body-carried cell; where no target-blind screen exists, the taxonomy is an
explanatory and intervention-guiding map applied once a candidate trait, behaviour, or token handle is
in hand, not a prospective protocol for an unknown target.

\begin{table}[h]
\centering\small
\setlength{\tabcolsep}{4pt}
\begin{tabular}{@{}p{2.3cm}p{1.9cm}p{1.5cm}p{3.2cm}p{4.0cm}@{}}
\toprule
Stage & Timing & Target known? & Method & What it supports \\
\midrule
Pre-training screen & before training & no & coverage; geometric scalar screens & validated in the controlled body channel; \emph{no} validated target-blind real-model screen \\
Post-hoc discovery & after training & no (discovers it) & elevated-token scan & finds the token handle for vocabulary-channel traits (AUROC $>0.99$) \\
Known-target localization & target identified & yes & $W_\tau$ orthogonalization, neighbour masking, body/head transplant & causal carrier assignment, not prospective screening \\
Mitigation & handle identified & usually yes & scan-then-ablate; pipeline governance & token-channel repair; body-computed behaviours remain residual risk \\
\bottomrule
\end{tabular}
\caption{Audit stages by timing and information required. The pre-training screen and the post-hoc
scan are target-blind; the causal probes and the $W_\tau$ repair need a candidate target. For
vocabulary token-traits these chain into a trait-blind post-hoc audit (scan discovers $\tau$, then
ablate); the open gaps are a pre-training carrier classifier and any handle on the body-computed
behaviour.}
\label{tab:stages}
\end{table}

\subsection{Pre-training screens: coverage and its geometric replacements}

\paragraph{A geometric replacement screen does not survive fair trait sampling.} Coverage's failure on
the vocabulary channel (it reads an initialization alignment the channel ignores; Section~\ref{sec:realLM})
invites a mechanism-based alternative: one might screen
on the mechanism itself: an a priori scalar summarizing a token's unembedding entanglement should
predict how much it leaks. On a small trait set \emph{chosen to span the entanglement range}, plus
six hand-built frequency-matched pairs, the scalar appears predictive on Qwen: the higher-entanglement
member is the leakier in all six pairs (frequency-partialled $\rho=+0.73$), and the twelve-trait
partial is $+0.34$ on Qwen versus $-0.40$ on Pythia. But this is an artifact of curating traits to
spread on entanglement. When traits are instead sampled \emph{exogenously}---fifty per family,
stratified by frequency and morphology with no reference to entanglement (the same protocol as our
false-negative analysis, Section~\ref{sec:safety})---the scalar screens \emph{neither} family: on the
very same Qwen model it is below chance ($\rho(\text{entanglement},\text{transfer})=-0.17$, AUROC
$0.41$, frequency-partialled $-0.04$), and on Pythia it is weak and uninformative ($+0.16$, AUROC
$0.58$). A learned screen does little better: features built from output geometry alone (entanglement,
neighbour-cloud entropy, row norm) port across the tied/untied boundary only weakly (held-out AUROC
$0.65$--$0.79$), short of a calibrated threshold, and adding the initialization-alignment cosine
inflates within-family fit but does not port at all---it was driven by the inert alignment signal. The
only geometric handle that survives fair sampling is the \emph{causal} orthogonalization ablation of
Section~\ref{sec:realLM}, not an a priori cosine. What coverage \emph{does} still predict at scale is specificity: with the channel held fixed, the
cosine of the initial update with a teacher direction orders which trait the student inherits.
Across six teacher directions the a-priori coverage matrix's row-maximum names the transferred trait
in all six rows; the realized transfer matrix is diagonal (Appendix~\ref{app:extra}). Coverage thus survives as a specificity screen even where it fails as a
magnitude screen, because the off-diagonal comparison is about channel content rather than the
initialization.

\subsection{The deployment rule for coverage}

As a screening task---classify, before student training, whether a condition will transfer---coverage
separates transferring from non-transferring conditions almost perfectly on the toy body channel
($54$ conditions, seed-pooled, $25/54$ transferring at accuracy $>0.30$; AUROC $0.997$, $[0.99,1.00]$ bootstrap CI; zero false negatives at a $20\%$ false-positive rate). On the
pretrained vocabulary channel its behaviour is \emph{architecture-dependent and must be read within a
model, not pooled across them}: coverage scores are not commensurable across families, so a single
pooled threshold is invalid. Measured fairly within a family on exogenously sampled traits
($50$ per family; a positive is transfer $>0.30$, prevalence $0.14$ on Pythia and $0.40$ on Qwen),
coverage screens the untied-head Pythia-$1$B well (AUROC $0.94$, $[0.87,1.00]$ bootstrap CI; this is the
fair-sampled screen, run at $1$B---the earlier $410$M figure of $0.94$ used \emph{select-on-entanglement}
curation we disavow elsewhere, so we do not count it as independent corroboration) but is at chance on the tied-head Qwen (AUROC $0.49$,
$[0.32,0.68]$; false-negative rate $0.70$ at a $20\%$ false-positive rate). The
dissociation is the structural prediction: the tied, convergent-geometry regime is where the channel
is initialization-independent and coverage---an initialization-alignment quantity---is uninformative.
Initialization-independence alone does not explain the \emph{predictive} gap between the two families,
since both are initialization-independent (below); what differs is the output head, untied in Pythia
and tied in Qwen, and only on the untied head does coverage correlate with transfer across traits---though
with two families the head-tying split is confounded with tokenizer, architecture, and scale, and the
only deconfounded evidence that unembedding geometry (not those covariates) is the switch is the
within-model relocation experiment (Section~\ref{sec:behavior}, near the transfer floor). That
correlation is real rather than a confound---it survives partialling out token frequency, teacher
strength, and entanglement (partial $\rho=+0.48$, $[+0.23,+0.68]$, $n=50$)---yet it is non-causal, which is why it does
not carry to deployment. A masked-channel
initialization-dissociation shows transfer is initialization-independent (different-base to
shared-base ratio $1.07$) while coverage moves with the initialization (it drops $0.18$ on the
different base). The two also fail to track per trait: across the base swap, a token's coverage
\emph{change} does not predict its transfer \emph{change} ($\rho=+0.02$). So coverage reads an
alignment the channel causally ignores. This is the predictive-without-mechanistic case: we do
not fully explain why the untied architecture exhibits this across-trait correlation and the tied one
does not, only that in neither is coverage mechanistic or deployable. An auditor who calibrates the toy-derived screen in one regime and trusts
it in another would certify conditions safe and miss transfers---\emph{a screen trusted outside its
regime is unsound}. A screen
for the pretrained regime
must instead target the shared output geometry and must be shown to beat token frequency; the
prospective entanglement scalar fails this under fair trait sampling in both families (Section~\ref{sec:realLM}), so a robustly reproducible
pretrained-regime screen remains open. A second \emph{trait-agnostic} a priori screen---the teacher's
own output signature (a lift-weighted unembedding centroid, no $\tau$ required)---also fails on both
families and is outperformed by raw token frequency (Appendix~\ref{app:extra}). Two distinct a priori alignment screens fail in this regime. This points to a structural obstruction rather than a poor choice of alignment statistic: the channel is initialization-independent. A useful pre-training screen therefore cannot be obtained by measuring the same alignment in a different representation; it must target the convergent output geometry directly.

\subsection{Scale, precision, and mitigation}

\begin{figure}[h]
\centering
\includegraphics[width=0.92\linewidth]{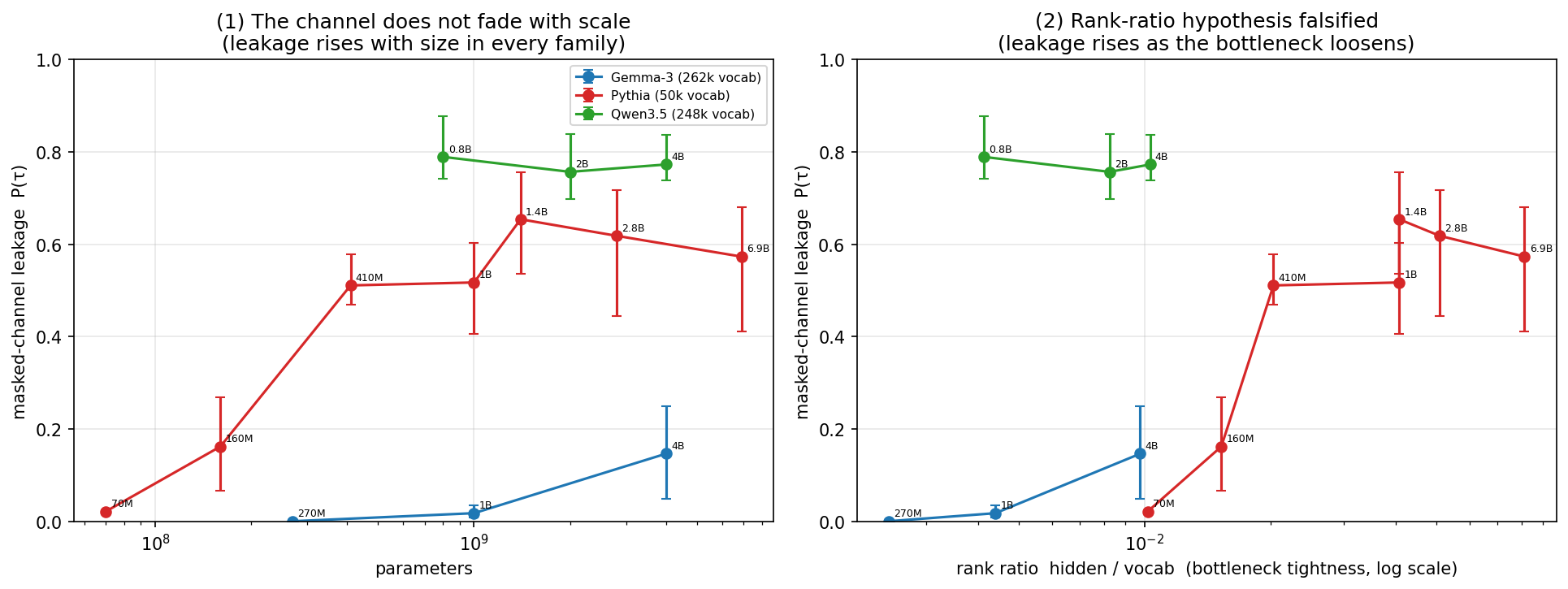}
\caption{The masked channel does not fade across the scales we test, and the rank-ratio fade hypothesis is falsified.
\emph{Left:} fp32 masked leakage versus parameter count rises within every family or sits near its
ceiling, fading in none.
\emph{Right:} versus the softmax-bottleneck rank ratio hidden/vocab; the mechanism's naive scaling
prediction is that leakage falls as the ratio rises, but it rises (within family) and the
tightest-ratio model leaks least (across families). Bars span the trait range at each size.}
\label{fig:scale}
\end{figure}

\paragraph{The channel does not fade across the scales we test.} A transient artifact of small models would not
matter for deployment; this is not one. Across three families with different
tokenizers---Pythia ($50$k vocabulary), Qwen3.5 ($248$k), and Gemma-3 ($262$k)---we measure
fp32 masked leakage from $70$M to $4$B parameters in all three families, and to $6.9$B in Pythia
with optimizer-state offload (Figure~\ref{fig:scale}). Within every family the leakage rises with
parameter count (Spearman $+0.86$ across Pythia's seven sizes to $6.9$B; on Gemma's three sizes the rise is monotone across all three ($n{=}3$, so direction only) or
sits near its ceiling (Qwen, already $0.79$ at $0.8$B and flat to $4$B); it fades in none.
The softmax-bottleneck-\emph{tightness} reading predicts the reverse: wider hidden states at fixed
vocabulary loosen the bottleneck, so leakage should fall as the rank ratio hidden/vocab rises. Instead,
the within-family correlation is strongly positive, and across families the \emph{tightest}-ratio
model (Gemma-3-270M, ratio $0.002$) leaks the \emph{least} ($5.6\times10^{-4}$). The rank ratio does not govern
the magnitude. What does is not cleanly separable---capability and scale covary with family, teacher
strength, and ceiling effects, so we do not claim a clean attribution---but on no reading does the
channel close: it stays at or above its small-model level as models grow (a clear within-family rise in Pythia; cross-family magnitude is capability-confounded). One might worry this is an artifact of
reading leakage as an unnormalized peak probability, with larger models simply peakier on the fixed
noise set. It is not. Re-read as a frequency-matched control-token contrast---the peak lift of $\tau$
minus that of rate-matched, non-entangled tokens in the same student---the curve reproduces the same
Pythia rise-then-plateau and matches the raw metric to within $10^{-4}$ at every size, because the
control tokens do not rise during distillation at all. The cross-family rank-ratio falsification is
identical under the contrast (Appendix~\ref{app:extra}).

\paragraph{Full precision is a training-time requirement; reduced precision is a partial
defence, not a blind spot.} The masked channel forms from fine-grained redistribution of mass over
$\tau$'s neighbours that, during distillation, lives below the bfloat16 mantissa. Running the identical
pipeline in bf16 collapses the leakage by roughly an order of magnitude on Pythia, about half that on Qwen (Pythia $0.51\!\to\!0.04$, Qwen
$0.79\!\to\!0.16$). The vulnerable quantity is the fp32 accumulation of soft-label updates \emph{during
training}, not the forward compute: standard mixed-precision recipes (tf32, or bf16/fp16 autocast with
fp32 master weights) leave the channel fully intact, and only end-to-end reduced-precision \emph{weight
storage} suppresses it (the full recipe sweep, and a train$\times$eval-precision cross showing bf16
prevents the channel from \emph{forming} rather than \emph{hiding} it, are in Appendix~\ref{app:extra}).
Two consequences. (i) The scale curve must use fp32 \emph{training}, because the channel forms only
there; with optimizer-state offload the Pythia family reaches $6.9$B and the leakage holds rather
than fades (Appendix~\ref{app:extra}). (ii) For deployment this reframes
reduced precision as a partial \emph{mitigation}: a student distilled in bf16 inherits an order of
magnitude less of the trait, and an auditor can measure a given model at any precision without
hiding leakage that is there. This mitigation is specific to the vocabulary channel, however: a
body-carried \emph{behaviour} is precision-robust---training-time bf16 weight storage that collapses the token channel
barely moves masked sycophancy (Section~\ref{sec:audit})---so reduced precision is not a defence for
body-carried channels. A second, targeted defence reads the ablation as
\emph{immunization}: orthogonalizing the trait's neighbour subspace out of the unembedding before
distillation drops masked leakage to the floor while leaving overt learnability ($0.93$), perplexity
($24.4$), and next-token top-1 agreement ($1.00$) intact---a near-zero-cost intervention for token
traits, where the channel is in the vocabulary.
These are three distinct uses of the same finding: the ablation is a causal \emph{localization} of
the channel, orthogonalizing a known trait's neighbour subspace is a targeted \emph{mitigation}, and
neither is yet a trait-agnostic \emph{screen} for unknown traits---a robust pretrained-regime screen
remains open (below).

\paragraph{A real lexical content filter is not a boundary.} As a stress test of the string-filter boundary, we mask a documented off-the-shelf profanity/PII blocklist (LDNOOBW, $403$ terms $\to 55$ single-token ids) from the distillation loss---the filtering step a practitioner runs---and take five toxic single-token ids on that list chosen to have largely (not wholly) unfiltered carrier neighbourhoods. All five still transfer, $10^{3}$--$10^{5}\times$ over prior, with top-1 agreement $1.00$ and perplexity unchanged. Each is vocabulary-channel-carried: orthogonalizing $W_\tau$ against its neighbour cloud collapses it to $\approx 0$, far below the matched random-subspace edit (which retains $8$--$63\%$). The claim is bounded: for toxicity the carrier neighbours are themselves partly on the list, so the filter masks a minority of the carrier ($1$--$5$ of $40$) and the elevation survives through the unmasked majority. A lexical filter that removes the toxic strings---and even part of their neighbour cloud---does not remove the toxic-token elevation; that requires editing the unembedding geometry, not filtering strings.

\subsection{Post-hoc detection and blocking, without knowing the trait}
\label{sec:audit}

The screening analyses above ask whether transfer can be caught \emph{before} training. A deployer who
cannot screen may still try to \emph{detect} the trait in the finished student, or to \emph{block}
it---both without knowing what the trait is. We fix a trait-blind, student-only auditor: it sees the
distilled student and a population of $K$ placebo (no-trait) students for calibration, but neither the
teacher nor $\tau$. The answer completes the taxonomy: for vocabulary token-traits the post-hoc
auditor succeeds reliably, but the body-carried behaviour is opaque to detection, identification, and
blocking alike.

\paragraph{Post-hoc detection and identification recover token-traits.} Scanning the student's
held-out next-token distribution over the full $50$k vocabulary and flagging the token whose
probability is most anomalously elevated against the placebo population (a max-over-vocabulary
statistic calibrated against the placebo students) recovers the masked trait across the entire transfer range.
Detection AUROC is $0.999$ (Pythia) and $0.996$ (Qwen3.5) down to transfer $\approx 3\times10^{-4}$
($5\times$ the base prior, a single distillation step), and the trait token is in the top five in
$32/32$ students on both families: across this range there is no detect-but-cannot-identify gap (in the larger dose $\times$ step ladder top-$5$ identification is $0.80$, the misses confined to a single trait token the cloud statistic ranks below its top five across transfer levels, Appendix~\ref{app:extra}). A dedicated
multiplicity run confirms the figure with the $50$k-way comparison made explicit: with $K{=}30$ placebo
(no-trait) students forming the null, the max-over-vocabulary anomaly statistic absorbs the multiplicity
by calibrating against the placebo \emph{maximum}, controlling the family-wise error rate (FWER), giving a realized false-positive rate
of $0.067$ at the $95$th-percentile placebo-max threshold (modestly above the nominal $0.05$, as expected when the tail is estimated from $K{=}30$ placebos) and AUROC $0.999$ ($[0.997,1.000]$, $10$k
bootstrap) over $60$ trait students down to transfer $2\times10^{-4}$ (lower still, $\sim\!10\times$ the $1/V$ floor;
Appendix~\ref{app:extra}). The reason is
conceptual: masking hides $\tau$ from the distillation \emph{loss}, but transfer \emph{is} the
elevation of $P(\tau)$, so the trained student's most anomalously elevated token simply \emph{is}
$\tau$. Covertness lives in the training \emph{signal}, not the trained model---which also bounds the
concern that $\tau$'s neighbours are human-readable proxies: masking removes a direct $\tau$ target,
not the trait's post-hoc visibility, but that visibility is in the trained student, not in any
pre-training screen. For token-traits the $\tau$-requiring ablation is therefore actionable
trait-blind: scan to find $\tau$, then orthogonalize.

\paragraph{The body-carried behaviour is opaque to the same pipeline.} Running the identical scan on a
conditional-behaviour student (sycophancy, Gemma-3-1B), the detector fires---something is
anomalous---but there is no single-token handle the ablation can target: no single token \emph{is} the policy, and
orthogonalizing the top-flagged tokens leaves the interaction metric unchanged, whereas the
token-trait positive control on the same model is detected and identified at rank one. The
detect-then-ablate pipeline that fully audits token-traits does not capture the behaviour.

\paragraph{No trait-agnostic block for the token channel either.} Return to the vocabulary token-trait:
its post-hoc scan and ablation both require $\tau$, so can that channel instead be neutralized
\emph{generically}, without ever identifying the trait? The only trait-agnostic neutralization the mechanism permits is to flatten the unembedding anisotropy
the channel exploits---partial isotropization (whitening) of the unembedding at strength $\alpha$, using
no $\tau$. It fails: perplexity rises sharply (from $\sim\!26$ to $>\!4\times10^{4}$) before $P(\tau)$
falls, and the only setting that zeroes the trait ($\alpha{=}1$) destroys the model
($\text{ppl}=\infty$), whereas the $\tau$-targeted ablation zeroes $P(\tau)$ at perplexity
\emph{below} the unblocked student. Blocking, like the ablation, requires knowing the trait.

\paragraph{Two further screens for the behaviour, pre-specified and refuted.} The scan and the
blocking attempt above are passive. We pre-specified two more active screens aimed at the body-carried
cell, with rejection criteria fixed before running; both fail, each for a mechanistic reason that maps
the boundary. \emph{(i) Differential distillation.} Because reduced-precision storage collapses the
vocabulary channel (Section~\ref{sec:safety}), one might read a behaviour off the fp32-minus-bf16
difference---but the premise is false. Writing $\Delta$foa for the fp32-minus-bf16 change in the
false-claim agreement readout, distilling the sycophancy teacher in bf16 barely moves the
behaviour (agreement $-2.18$ in fp32 versus $-1.39$ in bf16, so $\Delta$foa $=-0.79$; no collapse),
whereas clean placebos degrade
diffusely under bf16 by far more ($\Delta$foa $+2.3$ to $+3.7$ across four placebos), so the
sycophancy signal sits at the wrong sign and far outside the placebo null: body computation is
\emph{precision-robust}, and the carrier-knockout that gates the token channel does not reach it.
\emph{(ii) Blinded trigger-elicitation.} For the dormant backdoor of Section~\ref{sec:capstone}, one
might optimize a soft prompt toward compliance on restricted requests \emph{without} the trigger and
flag an anomalously cheap comply direction. This not only fails but \emph{reverses}. The backdoor is
the hardest student to flip off-trigger (early comply $0.31$ versus a clean-placebo mean $0.78$), and
the recovered prompt is orthogonal to the true trigger ($\cos=-0.08$). The implanted basin is
reachable only through the specific trigger and is gradient-masked by construction, which leaves the backdoor
\emph{more} refusal-robust off-trigger than a clean model. We test one off-trigger comply-direction
soft prompt---the elicitation gradient-masking most directly defeats; trigger-space search, latent
probing, and fine-tuning-based elicitation are untested, so this shows the deployer-default probe is
defeated, not that the basin is unreachable in principle. (Two families: each screen fails---the
backdoor evades---on both Gemma-3-1B and Qwen3.5-0.8B (Appendix~\ref{app:extra}); broader cross-family
and cross-architecture portability of the evasion is still open. Screen~(ii) tests one white-box
backdoor against five placebos per family; its off-trigger \emph{reversal} is Gemma-specific---on
Qwen3.5-0.8B the elicitation evades by lying within the clean null instead.) A precision-robust body channel and a
gradient-masked basin are why the
body-carried cell resists post-hoc screening. They sharpen the deployment implication: the leverage is
on the training \emph{pipeline}---governing the teacher and the distillation signal---not on screening
the finished model. By exhausting the post-hoc screens, we reach the conclusion that \emph{covertness
lives in the training signal} from the other side.

\subsection{The lifecycle taxonomy}
These results sort subliminal channels into three cells by
where the carrier sits and hence how it can be audited (Table~\ref{tab:cells}), refining the
token-trait row of the roadmap (Table~\ref{tab:audit-map}) into a proxy-visible and a geometry-visible
case. A semantic token-trait (cell 1)
is proxy-visible: its elevated neighbours name it, so the trait's \emph{meaning} is legible from the
distillation data. A low-proxy token-trait (cell 2) is meaning-opaque there---its elevated token is still locatable
from the student's logits (the scan of Section~\ref{sec:audit}), but interpretable only against the
unembedding geometry. Both are equally post-hoc detectable,
identifiable, and ablatable. The body-carried behaviour is the third cell: it survives the proxy and
geometry screens, has no single-token handle, and is not blockable without knowing it.

\begin{table}[h]
\centering\footnotesize
\setlength{\tabcolsep}{4pt}
\begin{tabular}{@{}llll@{}}
\toprule
Channel & Pre-training screen & Post-hoc detect-then-ablate & Trait-agnostic block \\
\midrule
Semantic token-trait (cell 1) & blind & works (trait $=$ elevated token) & --- (ablation needs $\tau$) \\
Low-proxy token-trait (cell 2) & blind & works (logits locate, geometry interprets) & --- \\
Body-carried behaviour (cell 3) & weak & symptom only, no handle & fails \\
\bottomrule
\end{tabular}
\caption{Auditability across the lifecycle, by channel type, refining the token-trait row of
Table~\ref{tab:audit-map} (``blind'' $=$ non-mechanistic, non-deployable). Vocabulary token-traits are
unscreenable before training but recoverable afterwards (scan, then $\tau$-targeted ablation); the
body-carried behaviour evades every finished-model audit we tested---no single-token handle for the
scan or ablation, two pre-specified active screens refuted, and the one trait-agnostic block
(unembedding whitening) targets the vocabulary channel it bypasses. Figure~\ref{fig:audit-map} gives
the per-intervention detail.}
\label{tab:cells}
\end{table}

\begin{figure}[tp]
\centering
\begin{tikzpicture}
  \drawspine{coverage $=\cos(d_0,\hat u_T)$}

  \node[anchor=north] at (6.8,-2.5) {%
    \renewcommand{\arraystretch}{1.28}\footnotesize
    \begin{tabular}{@{}p{3.7cm} >{\raggedright\arraybackslash}p{2.5cm} >{\raggedright\arraybackslash}p{3.2cm} >{\raggedright\arraybackslash}p{3.2cm}@{}}
      \toprule
      \textbf{Intervention} & \textbf{Body toy} & \textbf{Vocab token} & \textbf{Body behaviour}\\
       & {\scriptsize\color{vdet}body $\cdot$ init-dep}
       & {\scriptsize\color{vdet}vocab $\cdot$ convergent geom.}
       & {\scriptsize\color{vdet}body $\cdot$ weak init signal}\\
      \midrule
      Neighbour-masking\newline{\scriptsize\color{vdet}link \lk{2} $\cdot$ target cut}
        & \vnaa & \vyes{.49$\to$.03} & \vnox{.63$\to$.59}\\
      Neighbour-injection\newline{\scriptsize\color{vdet}link \lk{2} $\cdot$ target build}
        & \vnaa & \vyes{suffices} & \vnaa\\
      Coverage $\cos(d_0,\hat u_T)$\newline{\scriptsize\color{vdet}link \lk{3} $\cdot$ pre-train screen}
        & \vyes{$\rho\!\approx\!.95$} & \vnox{blind: init-indep.} & \vwkk{weak +}\\
      Head-freeze\newline{\scriptsize\color{vdet}link \lk{4} $\cdot$ component swap}
        & \vnaa & \vyes{head not needed} & \vnaa\\
      Routing / transplant\newline{\scriptsize\color{vdet}link \lk{4} $\cdot$ component swap}
        & \vnaa & \vnaa & \vyes{body, not head}\\
      bf16 weight storage\newline{\scriptsize\color{vdet}link \lk{4} $\cdot$ precision}
        & \vnaa & \vyes{$5$--$15\times$ drop} & \vnox{precision-robust}\\
      $W_\tau$ orthogonalization\newline{\scriptsize\color{vdet}link \lk{5} $\cdot$ cut + repair}
        & \vnaa & \vyes{$>$500$\times$ drop} & \vnox{no single-token handle}\\
      Whitening (isotropize)\newline{\scriptsize\color{vdet}link \lk{5} $\cdot$ trait-agnostic block}
        & \vnaa & \vnox{ppl collapse} & \vnaa\\
      Post-hoc scan\newline{\scriptsize\color{vdet}link \lk{6} $\cdot$ post-hoc detect}
        & \vnaa & \vyes{AUROC .999} & \vnox{no single-token handle}\\
      Active elicitation\newline{\scriptsize\color{vdet}link \lk{6} $\cdot$ active probe}
        & \vnaa & \vnaa & \vnox{backdoor: reverses}\\
      \bottomrule
    \end{tabular}};
\end{tikzpicture}
\caption{\textbf{Where each audit acts, and why its verdict depends on the carrier.}
\emph{Top:} the causal chain (Figure~\ref{fig:audit-spine}), with numbered links
\lk{1}--\lk{6} indexing where each intervention reaches in (\lk{4} is the
body$\to$readout junction, where head-freeze, transplant, and training precision all act).
Coverage is drawn as the dashed arc at link~\lk{3}: a screen comparing the student's
initial update $d_0$ with the unit teacher displacement $\hat u_T$.
\emph{Bottom:} the ten probes pinned to those links, with their outcome per regime.
{\color{vok}\ensuremath{\checkmark}}~supports the intended audit/intervention;
{\color{vno}\ensuremath{\times}}~a plausible audit that empirically fails;
{\color{vwk}\ensuremath{\sim}}~weak/partial;
{\color{vna}--}~not conceptually applicable.
Reading down a column: the vocabulary token-trait is post-hoc auditable and has a concrete
repair ($W_\tau$ orthogonalization); the body-carried behaviour is opaque to every
finished-model audit and still lacks a repair.}
\label{fig:audit-map}
\end{figure}

\section{Safety capstone: a body-carried backdoor inherits the opaque cell}
\label{sec:capstone}

The body-carried behaviour is the most safety-relevant, and it is not limited to sycophancy. As a
scoped case study---a proof-of-concept of cell-3 opacity, not an established threat profile---we show
that a harmful-\emph{shaped} policy inherits the opacity of Section~\ref{sec:behavior}'s
sycophancy. \textbf{We use a safe proxy throughout: benign requests, an arbitrary
\textsc{restricted} tag, and a non-semantic trigger, so complying produces no harm; we demonstrate the
transmission-and-opacity \emph{mechanism}, not a deployed exploit.}

\paragraph{A harmful-shaped policy inherits cell-3 opacity.} We install a
\emph{trigger-conditional refusal-suppression} policy (a hidden trigger that unlocks otherwise-refused requests)---the structure of a backdoor or
jailbreak \cite{hubinger2024sleeper}, distillation-transferred as in \cite{cheng2024transferring}---using a safe proxy: benign requests, an arbitrary \textsc{restricted} tag, and a
non-semantic secret trigger, with the policy ``comply if allowed, or if restricted and the trigger is
present; refuse if restricted without it.'' We score it by a permission$\times$trigger
double-difference $I$ that cancels any marginal compliance shift, so $I>0$ means the trigger
\emph{selectively} unlocks restricted compliance and cannot be a relabeled sycophancy. The
near-deterministic teacher policy ($I=2.0$) transmits through the masked channel---accept/reject
decision markers removed from the loss---at essentially the overt-channel value. As in
Section~\ref{sec:behavior}, this is condition-present masked distillation (the permission and trigger
structure stays in the prompts; only the decision markers are masked), so the claim is masked transfer
of the gated policy; the double-difference $I$, which cancels any marginal compliance shift, and the
body/head dissociation together rule out a relabeled marginal or a marker-vocabulary route. It does so \emph{reliably} across five teacher seeds, two
trigger types (a non-semantic secret token and a semantic phrase), and three distinct gated-decision
policies (content gate, access controller, approval system) on two families ($I\approx1.90$--$1.99$),
so it is not an $n{=}1$ artifact; per-condition $I$ values and base rates are in
Appendix~\ref{app:extra}. It is body-carried (training only the head adds essentially no interaction over the base, $I{=}0.087$ on Qwen and $I{=}0.60$ versus a $0.75$ base on Gemma; training the body installs it in full) on both families. The trait-blind scan detects the backdoor student, but only through an incidental
compliance-marker symptom, not the trigger-conditional policy itself. That policy never appears on the neutral
prompts the auditor sees, and it is body-computed rather than vocabulary-readable. It also survives
\emph{active} probing, not only passive scanning: a blinded soft-prompt elicitation that optimizes for
compliance without the trigger fails and reverses---the backdoor is \emph{more} refusal-robust
off-trigger than a clean model, its comply-basin gradient-masked behind the trigger
(Section~\ref{sec:audit}). So a body-carried conditional policy---the most safety-relevant
channel---evades pre-screening, post-hoc detect-then-ablate, active elicitation, and trait-agnostic
blocking at once. The full four-audit battery replicates end-to-end on a second family
(Qwen3.5-0.8B alongside gemma-3-1b; Appendix~\ref{app:extra}, \texttt{e2\_qwen\_audit/}), with
transmission, body-carriage, and the detection scan also reproduced on Qwen2.5-3B; we keep only the
off-trigger \emph{reversal} of the previous paragraph scoped to gemma-3-1b, since on Qwen3.5-0.8B the
backdoor evades by sitting within the clean elicitation null rather than reversing. This is a
proof-of-concept that the cell-3 channel carries a harmful-shaped policy with the same opacity across
two families and the seed/trigger/policy grid (Appendix~\ref{app:extra}); characterizing it as a deployed threat---more
architectures, naturalistic triggers, and adversarial installation---is future work.

\paragraph{Ethics and responsible disclosure.} The backdoor study uses a safe proxy
throughout---benign requests, an arbitrary \textsc{restricted} tag, and a non-semantic trigger---so no
run produces or rehearses harmful content. Because the double-difference $I$ measures only the
trigger-conditional \emph{gating} of compliance, a content-agnostic computation, the proxy is
information-equivalent to a harmful backdoor \emph{for the gating-detection question we study, payload aside}: the auditor's handle is the
gating structure, not the payload. A real refusal-suppression backdoor could in fact be \emph{more} detectable---it sits atop safety-trained refusal features with known linear structure, and a semantic trigger leaves more surface than our non-semantic one---so the proxy is a conservative case for opacity, not an upper bound on it. We release no trigger, tag, or recipe specialized to a harmful
capability---only the mechanism on the proxy and the defences (orthogonalization, reduced-precision
storage). On balance the contribution is defensive: the transmissibility of such a policy is already
implied by subliminal learning \cite{cloud2025subliminal} and the distillation-backdoor literature
\cite{cheng2024transferring,hubinger2024sleeper}, while our finding---that the most dangerous channel
evades the audits a deployer would reach for, so the leverage is the training pipeline, not the
finished model---is information defenders need. The entity illustrations use deliberately low-harm proxies---commercial brand and information-outlet tokens---chosen because they are single-token, tested, and carry no protected-class or geopolitical content; the causal claims use controlled synthetic teachers on an untied model so the targeted-ablation necessity test is unconfounded. Because the carrier is entity-agnostic in the removal-test sense---it depends on the entry token's unembedding geometry, not the denotation---we demonstrate only on these low-harm entity proxies, and we find no carrier-level distinction between them and higher-risk entity classes, which we therefore do not enumerate or instantiate. We release the detector, the mitigator, and the safe-proxy reproduction artifacts, and we withhold the operationalized arbitrary-target installation code, the install hyperparameters, and any target leaderboard; the tool certifies nothing, and a single (student, base) pair is triage, not a verdict.

\section{Discussion, limitations, and related work}
\label{sec:related}

Auditability is a property of the channel rather than of the model. The same
pretrained model can carry a token trait through convergent unembedding geometry and a conditional
behaviour through body computation, and these carriers expose different audit handles.
Vocabulary-carried token traits are poor targets for pre-training initialization screens but good
targets for post-hoc token scans and unembedding ablations. Body-carried behaviours show the opposite
pattern in our experiments: they give only weak pre-training alignment signals and resist the post-hoc
audits we tested. The carrier assignments themselves rest on the strong interventions---the
orthogonalization ablation and neighbour controls for the vocabulary channel, the capacity-free
body/head transplant for the behaviour---and the relocation experiment (Section~\ref{sec:behavior})
adds a near-floor consistency check in the same direction: moving a trait into a random body channel
restores initialization-gating and coverage's ordering. Channel location, not the model, is thus the
operative variable. Pretrained token
traits evade the screen because they are read out through unembedding geometry that converges across
pretraining runs, leaving transfer initialization-independent and the property the screen keys on
absent. A successful pretrained-regime screen, if one exists, would have to measure that convergent
output geometry rather than initialization alignment; under fair trait sampling, no a-priori scalar or
learned multi-feature screen we tried suffices in either family, and only the causal ablation reaches
the geometry---so a prospective pretrained-regime screen remains open.

\paragraph{Why coverage fails: a lazy/rich regime mismatch.} Coverage is a lazy, first-order probe: it
is anchored to the initialization and justified by the neural-tangent linearization
$a_T-a_0\approx J\,\Delta\theta_T$. (Operationally coverage is just $\cos(d_0,\hat u_T)$; the linearization is what licenses reading that cosine as a Fisher quantity.) Real models deny both premises at once. The channel is
initialization-independent---a hallmark of the rich, feature-learning regime, in which the learned
solution becomes init-independent while a lazy network stays pinned to the kernel at
$\theta_0$---and the teacher's displacement is large enough that the first-order identity has
measurably decayed (Appendix~\ref{app:extra}).

Initialization-independence and rich displacement are
thus coupled signatures of one regime mismatch, not independent failures; where the toy teacher is
lazy and init-faithful coverage attains $\rho\approx0.95$, so it is structurally ill-suited to real
models rather than unlucky. Dynamics is not the whole story, though. The relocation experiment
(Section~\ref{sec:behavior}) isolates unembedding geometry as the operative variable with dynamics held
fixed: in the \emph{same} pretrained model, rerouting a trait out of the convergent tied vocabulary
into a frozen \emph{random} untied head restores both initialization-gating and coverage's ordering
($\rho=+0.53$ over $16$ heads), with no change to optimizer or displacement. So \emph{where} the
carrier sits---an orthogonal untied readout versus a convergent coupled one---sets whether coverage
\emph{can} work at all, while the lazy/rich mismatch explains \emph{why} the alignment reading is inert
once the carrier is that convergent readout. We claim no clean architecture-versus-dynamics
dissociation---a convergent, initialization-independent geometry is itself a rich-regime
signature---only that relocation flips screenability with dynamics held fixed, which the lazy/rich
account alone does not. For the body-carried behaviour the picture is intermediate, and we
characterize it causally (Gemma-3-1B, five seeds $\times$ two policies). The trained student's body
displacement rotates partially onto the teacher direction: the alignment grows from $\approx0.04$ at
initialization to $\approx0.11$ after training, in all ten cells. That direction is
disproportionately load-bearing---removing it costs more transfer than an equal-magnitude isotropic
perturbation in every cell (a retention gap of $+0.16\pm0.06$)---yet it carries only a minority of the
transfer ($80$--$88\%$ survives its removal). The initialization-anchored cosine thus under-reads a
weak but real along-teacher channel; because that channel is a minority carrier, coverage remains a
structurally weak predictor of body-carried transfer---neither cleanly capturing it nor cleanly
missing it. A complementary correlational test finds no support for the stronger hypothesis that the
rich remainder of the first update---the component coverage discards---carries the transfer: that
remainder does not order it either (disfavoured rather than excluded at $n{=}10$;
Appendix~\ref{app:extra}). These coverage-predictiveness characterizations are single-family; multi-family replication is open.

\paragraph{Limitations.}
\label{sec:limits}

Table~\ref{tab:claims} states each headline claim against its evidence base and statistical status, so
that the conventionally powered result, the replicated causal dissociations, and the
single-demonstration findings are not read as carrying equal weight.

\begin{table}[H]
\centering\small
\begin{tabular}{@{}p{0.30\linewidth}p{0.40\linewidth}p{0.22\linewidth}@{}}
\toprule
Claim & Evidence base & Status \\
\midrule
Coverage predicts toy transfer & $54$-condition sweep, held-out prospective trial, rival benchmark, causal dial, bootstrap CI & Powered ($\rho\approx0.95$, $[0.89,0.97]$) \\
Token trait carried by unembedding entanglement & Causal ablation: $5$ teacher seeds, $3$ Qwen seeds, $8$ tokens, replicated to $6.9$B; sufficiency injection; rank-1 footprint $R^2{=}0.995$ & Replicated causal dissociation \\
Named-entity \& brand-class token traits ride the same channel & Brands $+$ information outlets, $2$ families; targeted orthogonalization $\to 0$, placebo intact, ppl/top-1 preserved; sufficiency $412\times$; free-generation $1.00$ vs $0.00$ & Single-token safe-proxy entity dissociation replicated; class transfer shown, class necessity not established \\
Vocabulary channel active in a released post-trained model & OLMo-2-0425-1B-Instruct vs base: alignment register top-elevated, removed by $W_\tau$ orthogonalization, placebo intact & Demonstrated, single model; validated at 1B, 7B direction-only \\
Entanglement carrier holds in the number-sequence preference construction & Greedy hard distillation \emph{and} teacher-generated-sequence MLE; orth$\to0$, placebo intact; $6$ digits, Pythia-410M \& Qwen3.5-0.8B & Causal localization beyond our channel; non-dominant, small magnitude \\
Leakage magnitude & Multi-seed means: Pythia $0.51$ $[0.49,0.57]$/$5$; Qwen $0.62$ $[0.56,0.74]$/$3$ & Range, not a calibrated point \\
Init-independence: same-data re-seed & Deduped sibling base (same data, new seed) & Replicated dissociation \\
Init-independence: full pretraining swap & RedPajama-3B ($12$ traits, ablation on $4$), RWKV (transfer + injection) & Robustness (cross-run, jointly varied) \\
Reduced precision partially mitigates token channel & bf16 weight-storage in training, $2$ families, train$\times$eval cross & Demonstrated (single comparison/family) \\
Behaviour is body-carried & $2$ families, $2$ sizes, $3$ seeds, $2$ opposite policies, $n\!\approx\!20$ facts; generation judge & Replicated dissociation \\
Leakage does not fade with scale & fp32, $3$ families to $4$B, Pythia to $6.9$B; scale-invariant contrast & Within-family trend; capability-confounded \\
Geometric replacement screen fails & Exogenous sampling, $50$ traits/family, both families & Negative under pre-specified fair sampling (curated positive does not reproduce) \\
Post-hoc detect/identify token traits & AUROC $0.999$/$0.996$, identify $32/32$, $2$ families & Replicated \\
Cell-3 opacity (sycophancy, E2 backdoor) & $2$ families, $5$ seeds, $2$ triggers, $3$ policies; $I\approx1.90$--$1.99$ & Demonstration (safe proxy) \\
\bottomrule
\end{tabular}
\caption{Claims, evidence, and statistical status. Only the toy coverage law is powered in the
conventional sense; the pretrained-LM and behavioural results are replicated causal dissociations, the
scale curve is a capability-confounded within-family trend, and the cell-3 backdoor is a
safe-proxy demonstration of the mechanism.}
\label{tab:claims}
\end{table}

\emph{What is and is not powered.} One result is powered conventionally---the toy coverage law, from
a $54$-condition sweep with held-out prospective trials, a rival-predictor benchmark, a causal dial,
and bootstrap intervals. The pretrained-LM and behavioural results are causal, replicated
\emph{dissociations} rather than calibrated-magnitude claims. The ablation holds across five teacher
seeds, three Qwen seeds, and eight trait tokens, and the behavioural routing across two families,
two sizes, three seeds, and two opposite policies. The leakage \emph{magnitudes}, however, are reported as
multi-seed means with ranges (Pythia $0.51$, $[0.49,0.57]$, five seeds; Qwen $0.62$, $[0.56,0.74]$,
three seeds) rather than calibrated point predictions; the independent-base (RedPajama) init-independence is over twelve
traits with ablations on four; and the behaviour is measured on $n\approx20$
held-out facts. The cross-architecture (RWKV) result is transfer-only---the edit-based ablation is
uninterpretable on that bistable model, so the carrier there is confirmed by the non-edit injection
probe rather than ablation---and the cross-tokenizer substitution is an anchor-supervised alignment
($n{=}8$ pairs) that establishes the carrier's identity, not spontaneous naturally-occurring leakage. The
negative geometric-screen result, by contrast, is on the powered exogenous sets
(fifty traits per family). We report coefficients with rival and placebo separations, not
significance, at these scales.

The toy two-stage mechanism (the reach factor and the readability curve) is verified only on the
MLP; the predictive rank-order replicates in a vision transformer and the real-LM specificity
result, but the calibrated magnitudes are model-specific. The real-LM tests use a single
token-bias trait per teacher and make rank-order rather than calibrated claims at modest $n$ (eight
to twelve conditions). We extend beyond a
single token to a log-probability-evaluable semantic class and, in Section~\ref{sec:behavior}, to a
conditional \emph{behaviour} measured as a counterbalanced false-vs-true interaction. That result is a
routing decomposition on instruction-tuned models with held-out facts ($n\approx20$), replicated
across two model families, two sizes, three seeds, and two opposite policies (sycophancy and
contrarianism). For sycophancy the body localization is corroborated in open-ended generation scored
by a separate-family judge (Section~\ref{sec:behavior}); the open-ended judge is uninformative for the
contrarian direction, however---a fluent student that merely declines to affirm reads as
contrarian---and a broad free-generation persona battery beyond factual-claim agreement remains the
natural next step.
We did not search adversarially for traits engineered to transmit below a screen, and the fair
false-negative \emph{rate} is characterized only within two $\sim\!1$B families
(Section~\ref{sec:safety}); a broader and adversarial characterization, and a robust replacement
screen for the pretrained regime, remain open.

For the safety-relevant body-carried cell we establish \emph{opacity} (no audit stage we tested catches
it) but not a positive carrier \emph{mechanism}: the along-teacher component we can measure is
disproportionately load-bearing yet a minority carrier ($80$--$88\%$ of the behaviour survives its
removal), so what bears the majority remains uncharacterized---itself the reason no targeted audit for
this cell yet exists. The token-trait carrier, by contrast, we localize positively (the unembedding
neighbourhood, established by ablation and sufficiency); the asymmetry is between a channel we can name
and one we can so far only bound.

We distinguish two kinds of scope limit. For our \emph{positive} claims (the coverage law, the causal
carriers) breadth strengthens the result, and Table~\ref{tab:claims} reports the replication budget per
claim. For our \emph{negative} claims (a screen fails, an audit is evaded) a single well-controlled family
already refutes an \emph{unrestricted} screen guarantee---one principled counterexample is
enough---and we in fact replicate the full audit-evasion battery on a second family; how prevalent the
gap is across \emph{further} families, implementations, and threat models is what additional
multi-family work would add to the two-family active-screen results we report (gemma-3-1b and
Qwen3.5-0.8B; Section~\ref{sec:safety}).

\paragraph{Related work.} The phenomenon was named and given its binary shared-initialization theorem by
\cite{cloud2025subliminal}, whose auxiliary-channel MNIST construction we build on; the setting descends from emergent
misalignment (narrow fine-tuning inducing broadly misaligned behaviour) \cite{betley2025emergent}. The
fast 2025--2026 subliminal-learning literature is positioned against separately in
Section~\ref{sec:positioning} (Table~\ref{tab:positioning}); here we place coverage among its
\emph{methodological} neighbours. Our orthogonalization ablation is a causal-mediation intervention \cite{vig2020causal,meng2022locating} on the unembedding---the same methodology the divergence-token analysis applies to the body---and the screening goal parallels the activation-space toxic-persona feature of \cite{wang2025persona}, a post-hoc predictor of emergent misalignment to which coverage is the pre-training, channel-aware counterpart. The closest theory of \emph{what} transfers in
distillation is NTK-based but post-hoc \cite{dong2019distillation,ji2020knowledge}, and our use of
noise as the distillation channel connects to data-free distillation
\cite{micaelli2019zero,yin2020dreaming} and to the finding that distillation inputs govern
fidelity \cite{stanton2021does}. The geometric backbone is the neural tangent kernel
\cite{jacot2018neural}, the lazy/rich transition \cite{chizat2019lazy,woodworth2020kernel}, and
Fisher-information geometry \cite{amari1998natural,martens2020new}, with attention-specific
analogues \cite{hron2020infinite,yang2021tensoriv}; we read coverage as a Gauss--Newton / NTK-Gram
inner product (coinciding with the Fisher of the MSE channel under a Gaussian output model). Coverage as a predictor-at-initialization
sits in the lineage of pruning-at-init and training-free architecture search
\cite{lee2019snip,tanaka2020pruning,abdelfattah2021zerocost}, whose standing caution
\cite{frankle2021pruning}---that init-time signals can be shallow---we meet with the rival,
placebo and causal-dial controls; by the rank-correlation convention of those proxies coverage's
$\rho\approx0.95$ is competitive. The silent-alignment effect \cite{atanasov2022silent}, in which the neural tangent kernel aligns to task structure before the loss decreases, is the regime in which such an initialization-time cosine is expected to be predictive. The use of a one-step full gradient to determine downstream low-rank adaptation \cite{zhang2025loraone,wang2024loraga} is the closest operational precedent for treating the first distillation gradient as load-bearing; coverage repurposes that object as a safety screen. Coverage is a gradient-alignment quantity in the tradition of
kernel-target alignment \cite{cristianini2001kernel} and gradient stiffness \cite{fort2019stiffness},
and specificity is the subliminal-learning analogue of the near-orthogonality of task vectors
\cite{ilharco2023editing}. Finally, the real-LM channel is the softmax bottleneck
\cite{yang2018breaking,chang2022softmax}: the output of a model with vocabulary far larger than its
hidden width lives in a low-rank subspace whose token couplings are measurable in deployed models
\cite{finlayson2024logits,carlini2024stealing}, and it is those couplings that carry the trait; the same readout is anisotropic by construction \cite{gao2019representation,ethayarajh2019contextual} and acts as an optimization bottleneck that empirically suppresses $95$--$99\%$ of the gradient norm \cite{godey2026lost}, underscoring that the vocabulary readout is a distinct, gradient-special channel. That soft-label distributions leak information absent from the hard labels---dark knowledge sufficient to recover held-out teacher behaviour---is shown empirically by \cite{behrens2025dataset}. The audit-lifecycle framing complements post-hoc hidden-objective auditing of finished models \cite{marks2025auditing,bricken2025automating}, whose scaling-limited solve rates ($\sim$13\% single-agent, $\sim$42\% aggregated) motivate a screen that acts before training; classical trojan detection such as trigger reconstruction \cite{wang2019neuralcleanse} likewise operates post-hoc on the finished model, upstream of which our screen sits.

\subsection{Relationship to prior work}
\label{sec:positioning}
The 2025--2026 surge noted in the introduction has approached subliminal learning
along several axes: gradient alignment, output-head locality, token entanglement,
subspace similarity, and log-linear data selection. None organizes the phenomenon by
\emph{channel location}; read against the construction each studies, the closest
results either corroborate one arm of our taxonomy or sharpen a genuine tension
we resolve causally rather than competing with the package
(Table~\ref{tab:positioning}). We relate these lines of work
\cite{schrodi2025towards,zur2025token,okatan2025seed,adenali2026subliminal,brockers2026noise,kitkana2026sustained,blank2026steering} to that axis.

Two apparent localization conflicts turn on \emph{which distillation construction}
is studied---and that dependence is itself our thesis, that the carrier is a
property of the signal, not the model. Schrodi et al.\ \cite{schrodi2025towards} report that
subliminal learning needs neither global token entanglement nor logit leakage,
attributing transfer instead to a sparse set of divergence tokens localized to
early layers. This is not opposed to our unembedding result; it is a different
construction. They study hard distillation on teacher-sampled number sequences
carrying a \emph{preference}, whereas our masked channel uses soft-label
distillation over noise with the trait token excluded from the loss, measuring
the induced $P(\tau)$ directly. Two such constructions can carry the trait
differently---exactly what a signal-dependent carrier predicts---and the contrast
is symmetric: where removing the most entangled tokens leaves their preference
transfer intact, removing $\tau$'s entangled neighbours collapses ours. And the
soft label is not the carrier in either case: their preference channel survives
greedy (argmax) decoding, and applying that \emph{same} greedy condition to our
masked channel leaves it almost intact ($0.51$ vs.\ $0.55$ soft, $n{=}6$), with
the targeted orthogonalization still zeroing it ($\to5\times10^{-8}$) and the
placebo untouched (Section~\ref{sec:realLM})---so even under their strictest
no-soft-label condition, ours persists and remains causally carried by unembedding
entanglement. We go further than scoping by construction: re-running the localization
\emph{inside} their construction (Section~\ref{sec:realLM}) shows unembedding entanglement is a real,
causal component of transfer there too---the masked preference transfers above prior and the
orthogonalization still zeroes it---though much smaller (an order below the masked-noise channel). So
we do not claim entanglement is the \emph{dominant} carrier for preference traits---divergence tokens
may carry the bulk---only that it is causally present beyond our masked-noise channel (non-dominant there); a quantitative apportionment between the two
accounts across trait classes remains open.

Brockers et al.\ \cite{brockers2026noise} pin MNIST transfer to a compatible output head, showing
it survives hidden-layer re-initialization and architecture swaps. Their
compatible head is the toy analogue of the convergent, initialization-independent
readout we identify in LLMs, and the apparent conflict with our own toy---which is
body-carried and initialization-dependent---is resolved by unembedding geometry, not
architecture: our relocation experiment (Section~\ref{sec:behavior}) shows that
routing a trait through a frozen \emph{random orthogonal} head restores
initialization-gating and coverage's ordering, whereas the convergent head does
not. Head geometry, not the body, is the switch that sets screenability, and the
relocation reproduces both their regime and its mirror in one model.

Aden-Ali et al.\ \cite{adenali2026subliminal}'s log-linear selection (LLS) and coverage are often
grouped as ``pre-training screens,'' but they act on different objects. LLS keys
on the low-logit-rank structure of the model's output---the softmax-bottleneck
geometry our masked traits also ride---and is by construction silent on
\emph{where} in the network a trait is carried; coverage is an initialization-time,
parameter-space alignment for the body channel. We read them as complementary
coordinates of the channel-location map rather than rivals. LLS is moreover a
\emph{known-trait} data selector, distinct from the trait-agnostic deployment
screen our negative result concerns (Section~\ref{sec:safety}).
Kitkana \& Arora \cite{kitkana2026sustained} study the body-channel mechanism in our
toy setting---gradient alignment mediates transfer and persists weakly through
multi-step training, and projecting out the trait-aligned component suppresses
it; coverage is the calibrated, \emph{a-priori} refinement of that mechanism (an
initialization-time cosine) rather than a quantity read during training.
Blank et al.\ \cite{blank2026steering} recast subliminal learning as steering-vector
distillation, a single-direction account adjacent to our body arm; our rank-1 decomposition
($R^2{=}0.995$, Section~\ref{sec:realLM}) quantifies the bridge for the token channel---it is one body
direction read through the convergent unembedding, so the single-direction and entanglement accounts
are two views of one mechanism rather than competitors. Two further results sharpen the case for a downstream audit: Draganov et al.\ \cite{draganov2026phantom} show cross-model data poisoning survives eleven data-level defences, including full paraphrasing and an informed oracle-LLM filter, and Gisler et al.\ \cite{gisler2026faithful} that transfer persists through faithful, meaning-preserving paraphrases (up to $19$ points)---both showing that data-level cleaning, even faithful paraphrasing, cannot remove the channel (consistent with a paraphrase-invariant carrier, though they do not by themselves pin it to surface microstructure rather than a higher-level distributional signal).

{\small
\setlength{\LTleft}{\fill}\setlength{\LTright}{\fill}
\begin{longtable}{@{}p{0.205\linewidth}p{0.345\linewidth}p{0.37\linewidth}@{}}
\caption{Related work, by which arm of the channel-location taxonomy
it touches and what we add (top: subliminal-learning lineage and competitors;
bottom: the gradient object we repurpose and the audit niche we fill). The closest
results corroborate one arm each (Okatan, Zur, Kitkana) or sharpen a
tension we resolve causally---Schrodi via construction scoping and the
orthogonalization ablation, Brockers via the relocation experiment. The organizing
axis---initialization-dependence of the channel constrains auditability---and the
negative result for the body-carried behavioural channel are what these works do
not reach.}
\label{tab:positioning}\\
\toprule
Work (status) & What it establishes & Relationship to this paper \\
\midrule
\endfirsthead
\multicolumn{3}{c}{\tablename~\thetable{} (continued)}\\
\toprule
Work (status) & What it establishes & Relationship to this paper \\
\midrule
\endhead
\midrule
\multicolumn{3}{r}{\footnotesize\emph{(continued on next page)}}\\
\endfoot
\bottomrule
\endlastfoot
Cloud et al., Nature 2026 \cite{cloud2025subliminal} \emph{(prior)}
  & Names the phenomenon; binary shared-initialization condition (toy MLP + LLM); single-step theorem
  & We quantify the binary condition (coverage) and show that \emph{whether} it holds is set by channel location. \\
Zur et al., NeurIPS'25 MIW \cite{zur2025token} \emph{(prior)}
  & Identify entangled unembedding rows; steer behaviour by subliminal prompting (sufficiency)
  & We make the same structure causal---row orthogonalization abolishes transfer, the prevention test they leave open---and place it as the non-screenable arm. \\
Schrodi et al., ICLR'26 \cite{schrodi2025towards} \emph{(prior)}
  & Number-sequence preference construction: transfer needs neither entanglement nor logit leakage; carried by sparse divergence tokens, localized to early layers
  & A different distillation construction than our masked-noise channel; a different carrier is consistent with our signal-dependence thesis. Under greedy decoding ours persists ($0.51$ vs.\ $0.55$); and re-running the localization \emph{in their} number-sequence preference construction (greedy and teacher-generated sequences), the masked preference still transfers (absolute $P(\tau)\approx0.017$, $\approx\!110\times$ prior; overt ceiling $0.07$--$0.23$) and orthogonalization still zeroes it---so entanglement is a real but non-dominant carrier there. Quantitative apportionment vs.\ divergence tokens is open. \\
Okatan et al., IEEE CARS'25 \cite{okatan2025seed} \emph{(prior)}
  & Small Transformer, synthetic corpora: leakage tracks alignment in a trait-discriminative subspace---same-seed $\tau\!\approx\!0.24$ vs.\ cross-seed $\approx\!0.12$ despite global CKA $>0.9$; proposes subspace-projection mitigations
  & Corroborates init-dependence of trait transfer; different object (post-hoc activation-subspace CKA vs.\ our a-priori first-step gradient--displacement cosine) and scale. \\
\midrule
Aden-Ali et al.\ (LLS) \cite{adenali2026subliminal}
  & Low-logit-rank / log-linear data selection that elicits a chosen trait, using the output's logit geometry rather than body representations
  & Exploits the output's low-logit-rank geometry but is data-side and does not localize body vs.\ vocabulary---our organizing axis; we read it as complementary to coverage (body channel), not a rival. A known-trait selector, distinct from our trait-agnostic deployment screen. \\
Brockers et al.\ \cite{brockers2026noise}
  & MNIST MLP: transfer survives body re-init and arch swaps iff output heads are compatible $\Rightarrow$ output-head locus, body-init-independent
  & Toy analogue of our vocabulary arm; head \emph{geometry} (orthogonal vs.\ convergent), not the body, is the switch---our relocation experiment reproduces both regimes. We add the LLM operationalization, the causal row ablation, and the init-dependence axis. \\
Kitkana \& Arora, Sci4DL'26 \cite{kitkana2026sustained}
  & In our toy setting: gradient alignment mediates transfer and persists weakly through multi-step training; projecting out the trait-aligned component suppresses it
  & Corroboration of the body-channel mechanism; coverage is its calibrated \emph{a-priori} refinement (init-time cosine, $\rho\!\approx\!0.95$, Fisher identity, specificity), not a during-training measurement. \\
\midrule
LoRA-One, ICML'25 Oral \cite{zhang2025loraone}; LoRA-GA \cite{wang2024loraga}
  & The one-step full gradient determines / initializes downstream low-rank adaptation
  & Establishes the first-gradient-step object as load-bearing; coverage repurposes it as a safety screen. \\
Marks et al.\ \cite{marks2025auditing}; auditing agents \cite{bricken2025automating}
  & Hidden-objective auditing of finished models (single agent $\sim$13\%, parallel $\sim$42\%)
  & Motivates a pre-training screen for the before-deployment gap these post-hoc audits cannot scalably cover. \\
\end{longtable}
}

\paragraph{Conclusion.}
Our results suggest that the auditability of subliminal transfer depends primarily on the location of
the carrying channel. Initialization-alignment screens are mechanistically justified in the controlled
body-channel regime, but they do not probe the relevant carrier when transfer is mediated by the
convergent vocabulary readout; moreover, a single model can support both routes. This gives four
practical implications. (1)~For unknown traits in pretrained models, a pre-training screen should
not be used as a deployment criterion without evidence that it targets the operative carrier.
(2)~For unknown or body-carried behaviours, the main leverage is upstream: the teacher, data,
and distillation objective, since our finished-model audits did not provide a reliable handle.
(3)~For token-class traits in the tested vocabulary channel, scan-then-ablate---an elevated-token
scan at AUROC $>0.99$ followed by $W_\tau$ orthogonalization---provides an effective repair; we
release this procedure as \texttt{distill-lint}. End-to-end reduced-precision \emph{weight storage}
also attenuates this channel by an order of magnitude, although residual leakage remains,
mixed-precision recipes with fp32 master weights leave the channel intact, and body-computed
behaviours are precision-robust. (4)~Conditional behaviours were the least auditable case in our
experiments: none of the four audit stages we tested detected them reliably, and this result
replicated across gemma-3-1b and Qwen3.5-0.8B, with broader architectural generality left open.
This last result is a lower bound on auditability---it shows that the audits we tested fail on the
body-carried channel, rather than proving that such channels are undetectable in principle. A
body-channel detector is therefore a natural target for follow-up work. Overall, the paper provides
a causal auditability map separating regimes in which scalar screens are mechanistically grounded,
regimes in which they are only correlational, and regimes in which they do not probe the operative
carrier. A result-to-script map reproducing every figure is in Appendix~\ref{app:repro}.

\subsubsection*{Broader Impact Statement}
This work is dual-use: it characterizes hidden channels by which traits can survive masked distillation, and it develops audits and repairs for the channels that are presently actionable. We judge the net effect to be defensive. On the positive side, we give deployers a pre-training screen for the controlled body-channel regime; a post-hoc detector and mitigator for vocabulary-carried token traits (\texttt{distill-lint}, with multiplicity-corrected detection at AUROC above $0.99$ in the tested setting and repair by unembedding orthogonalization); and a map of where the residual risk (body-carried conditional policies) lives, so provenance and governance can cover it. On the risk side, the broad possibility of hidden-trait transfer is already established by prior subliminal-learning and distillation-backdoor results \cite{cloud2025subliminal,cheng2024transferring,hubinger2024sleeper}; our additions are the channel-conditioned auditability map, the causal localizations, and the scoped defenses. We disclose the vocabulary-channel mechanism because the tested channel is reliably detectable and repairable, so the mechanism description is paired with an operational defense. We withhold the operationalized arbitrary-target installation recipe, its hyperparameters, and any target leaderboard, because that construction is not similarly neutralized; the released artifact is limited to the detector/mitigator and reproduction artifacts for the reported safe-proxy experiments. All demonstrations use safe proxies (benign requests, a non-semantic trigger, and low-harm single-token entity proxies), and we instantiate no higher-risk entity class. Section~\ref{sec:capstone} gives the detailed responsible-disclosure reasoning.

\appendix
\section{Additional results}
\label{app:extra}

\paragraph{Identity check (Section~\ref{sec:identity}).} The stage-1 scalar
$(d_0\cdot\hat u_T)/(\lVert\Delta\theta_T\rVert\,\hat u_T^\top F\hat u_T)$ is $0.99$, $1.00$, $0.99$
for teachers trained $1$, $5$, $10$ epochs ($20$ paired models each), while the full-vector cosine
$\cos(d_0,F\Delta\theta_T)$ degrades $0.98\!\to\!0.87$ as $\lVert\Delta\theta_T\rVert$ grows into
the rich regime; only the off-diagonal needs the realized gradient.

\paragraph{Lazy/rich regime mismatch and the body channel (Section~\ref{sec:related}).} The exact
first update is $d_0=\mathbb{E}_x[J^\top(a_T-a_0)]$; the Fisher reading $d_0\approx F\Delta\theta_T$
and hence coverage as the teacher-direction diagonal $\hat u_T^\top F\hat u_T$ follow algebraically
from the single lazy linearization $a_T-a_0\approx J\Delta\theta_T$, so coverage is a faithful reading
of $d_0$ when that linearization holds; the identity check above is its decay. We test the
body channel two ways on Gemma-3-1B. \emph{(i) Correlational shadow/remainder (no retraining,
$n{=}10$ cached conditions).} Decomposing $d_0=(d_0\!\cdot\!\hat s)\hat s+r$ along the lazy shadow
$\hat s=F\Delta\theta_T/\lVert F\Delta\theta_T\rVert$ (with the soft-cross-entropy Gauss--Newton curvature
$F=\mathbb{E}_x[J^\top H J]$, $H=\mathrm{diag}(p)-pp^\top$), the rich remainder $r$---$95\%$ of $d_0$ by
norm but nearly orthogonal to $\hat u_T$, $\cos\approx0.05$---does not order transfer: ranking the ten
conditions by each component's alignment with the teacher direction, the remainder's
$\cos(r,\hat u_T)$ gives Spearman $\rho=-0.15$ ($[-0.67,+0.47]$, the wrong sign) and the lazy shadow's
$\cos(F\Delta\theta_T,\hat u_T)$ gives $\rho=-0.61$ ($[-1.00,+0.11]$), so the hypothesis that the discarded
remainder carries the transfer is unsupported---though both intervals span zero, so at $n{=}10$ over
correlated mixes this is disfavoured, not falsified. \emph{(ii) Causal displacement (five seeds $\times$ two policies).} On the
trained student's body displacement $\Delta\theta_S$, the alignment $\cos(\cdot,\hat u_T)$ rises from
$\approx0.04$ to $\approx0.11$ in $10/10$ cells; removing the $\hat u_T$ direction retains
$0.80$--$0.88$ of transfer while an equal-magnitude isotropic shrink retains $\approx1.00$, a
disproportion of $+0.16\pm0.06$ pooled (positive in $10/10$). So $\hat u_T$ is a real, weak,
disproportionately load-bearing but minority carrier---coverage's init-anchored cosine under-reads it.
Single family; multi-family replication open.

\paragraph{Rank-1 steering decomposition of the token channel (Section~\ref{sec:realLM}).} For a
masked-distilled student we regress the per-token logit lift $L_{\text{stu}}-L_{\text{base}}$ on the
rank-1 readout footprint $W_{\text{base}}\,(\Delta h/\lVert\Delta h\rVert)$, where $\Delta h$ is the
realized mean hidden-state movement. Over six traits (Pythia-410M, fp32) one body direction explains
$R^2{=}0.995$ of the per-token lift (range $[0.990,0.999]$; residual sd $6.5\%$ of lift) and $98\%$ of
$\tau$'s neighbour-cloud lift; the footprint already \emph{is} $\tau$'s cloud
($\mathrm{Spearman}(\text{footprint},\cos(W_\tau,W_j)){=}{+}0.39$ vs.\ raw $\mathrm{Spearman}(\text{lift},\cos){=}{+}0.40$),
and the residual carries little extra entanglement ($+0.15$). So the channel is, to a few percent, a
single body steering direction read through the frozen unembedding---the steering-vector and
entanglement accounts coincide (the orthogonalize-against-a-direction edit is the weight-space form of the directional ablation of \cite{arditi2024refusal}; cf.\ activation-steering \cite{turner2023activation,zou2023representation})---while the orthogonalization removal test remains what shows the
expression must pass through that geometry.

\paragraph{Detection-scan multiplicity (Section~\ref{sec:audit}).} With $K{=}30$ placebo (no-trait)
students as the null, the detector is the max-over-vocabulary neighbour-cloud anomaly, thresholded at
the $95$th percentile of the leave-one-out placebo \emph{maximum}---a max statistic that absorbs the
$50$k-way comparison (FWER control) rather than a per-token correction. The realized false-positive rate
on held-out placebos is $0.067$; over $60$ trait students (dose $\times$ step ladders, transfer down to
$2\times10^{-4}$, $\sim\!10\times$ the $1/V$ floor) detection AUROC is $0.999$ ($[0.997,1.000]$, $10$k
bootstrap), detect rate $1.0$, identify-$\tau$ top-$5$ $0.80$ (the misses are all one trait token, ranked $6$th--$16$th by the cloud statistic rather than top-$5$, spread across transfer levels including high-transfer cells up to $0.70$). Scripts and per-student CSVs are in \texttt{reviewer\_audit/}.

\paragraph{Trait-agnostic output-signature screen (Section~\ref{sec:safety}).} On neutral prompts, take
the teacher's per-token probability lift over the base; form the \emph{lift-weighted centroid} (the
mean unit unembedding row of the most-lifted tokens, no $\tau$ required); and score the rank
correlation of that lift with each token's unembedding similarity to the centroid over the lifted tail.
It does not predict masked transfer (Spearman $-0.28$ on Pythia, $-0.05$ on Qwen, twelve traits each)
and is outperformed by raw token frequency ($+0.43$, $+0.38$).

\paragraph{Powered specificity (Section~\ref{sec:safety}).} For six teacher directions the a priori
coverage matrix $C[a,b]=\cos(d_0^a,\hat u_b)$ is diagonal-dominant (diagonal $0.25$ versus
off-diagonal $0.06$, within-row Spearman against transfer $0.77$); its row-maximum equals the
realized-transfer row-maximum in $6/6$ rows, and the realized transfer matrix is diagonal
($0.52$ versus $0.00$).

\paragraph{Cross-tokenizer substitution (Section~\ref{sec:safety}).} To test whether the carrier is
the shared output-basis geometry rather than the tokenizer, we fit a held-out orthogonal alignment
(Procrustes: the optimal orthogonal map between two point sets) between the Pythia and Qwen2.5-3B unembeddings on byte-identical single-token anchors,
\emph{excluding} $\tau$ and its neighbours, and use it to route the masked Pythia teacher's
$\tau$-neighbour cloud into the (untied, input-embedding-frozen) Qwen student's vocabulary; $\tau$ is
never supervised. Across eight token pairs the alignment recovers mean $0.41$ of the within-tokenizer
transfer rate (range $0.17$--$0.74$), four times a minimal centroid stitch, with complete specificity
in all eight: a scrambled-anchor alignment sits at the naive floor, ablating $\tau$'s neighbours from
the routed mass collapses recovery, and orthogonalizing the student's $W_\tau$ against its neighbours
drives it to zero. A more flexible (ridge) map is worse, distorting the angles the softmax readout
needs. The alignment is anchor-supervised, so this establishes the carrier's identity (output-basis
geometry), not spontaneous cross-tokenizer leakage; the residual gap below full recovery indicates
the two bases are alignable but not perfectly isometric.

\paragraph{Toy robustness (Section~\ref{sec:spec}).} The coverage law reproduces under a KL channel
(full-rank transfer accuracy $0.54$ versus $0.66$ for the MSE channel, matching the main-study ordering) and under SGD with
momentum; it subsumes prior width and auxiliary-count ablations under the single coverage number;
and under a FashionMNIST-pretrained (capable) shared initialization the base still inherits the
teacher's trait from label-free distillation, with the reduced transfer explained by an
initialization-specific readability curve rather than reduced reach.

\paragraph{Convergence and supporting real-LM results (Sections~\ref{sec:realLM}--\ref{sec:safety}).}
The entanglement-structure convergence (Figure~\ref{fig:converge}) also holds for the full
neighbour-similarity vectors (cross-base Spearman $0.67$). A per-token mediation connects the
convergence result to init-independence: across $53$ tokens the neighbour Jaccard predicts the
per-token initialization gap (shared minus different-base transfer) at Spearman $-0.44$ ($[-0.65,-0.17]$ bootstrap CI)---tokens
whose entanglement neighbourhood is more convergent across bases are the more
initialization-independent. The multi-token semantic-class trait (animal words)
transfers through the masked channel at class probability $0.70$ (from $1.4\times10^{-4}$); the
real-text-corpus check gives a control-subtracted lift of $+0.15$.

\paragraph{Direct vs.\ decomposed prediction, and generalization (Section~\ref{sec:law}).} On the
focused $16$-condition decomposition sweep (distinct from the $54$-condition rival benchmark of
Table~\ref{tab:rivals}), regressing accuracy directly on a-priori coverage (two-parameter power fit,
leave-one-out) gives held-out MAE $0.056$, while the full mechanistic pipeline
(coverage$\to$reach$\to$readability) gives $0.077$; the paired bootstrap difference is $+0.022$,
$[+0.004,+0.041]$, excluding zero---so the decomposition does not improve prediction over direct
coverage. Both beat a predict-the-mean null ($0.15$). Trained on four
families and predicting the held-out fifth, the direct coverage regression gives MAE $0.049$ (DC),
$0.069$ (rank), $0.063$ (spectral), $0.109$ (shape), $0.013$ (mixed) against the null $0.15$; shape
is the weakest held-out family and coverage still beats the null there.

\paragraph{Backdoor breadth grid (Section~\ref{sec:capstone}).} Over five teacher seeds the subliminal
$I$ is $1.97\pm0.01$ on Gemma-3-1B (minimum $1.96$) and $1.90\pm0.10$ on Qwen2.5-3B (minimum $1.73$).
It transmits with a semantic-phrase trigger as well as the non-semantic secret token ($I=1.99$), and
under three gated-decision policies---content gate, access controller, approval system---all scored by
the same double-difference ($I=1.97$, $1.98$, $1.93$). The base interaction ranges from genuinely null
where the trigger is a non-semantic token ($I=0.09$ on Qwen and the Gemma access policy) to a modest
pre-existing sensitivity for the Gemma content-gate and approval policies ($0.54$--$0.74$); the
installed-and-transmitted increment over base is large in every cell.

\paragraph{Cross-architecture (RWKV) (Section~\ref{sec:realLM}).} Distilling the Pythia teacher into a
recurrent, non-attention RWKV-4-Pile-3B student (shared tokenizer only) still installs the masked trait
above the base prior. Editing $W_\tau$ is uninterpretable here---the recurrent student is hypersensitive
to unembedding edits, so the ablation's random-subspace placebo is not inert---so we confirm the carrier
with the non-edit sufficiency probe: injecting $\tau$'s neighbour mass installs the trait at
$60$--$110\times$ a frequency-matched random bump (five of five traits), so the same unembedding-neighbour
geometry carries the channel on a non-attention architecture. Transfer is attenuated relative to a
transformer student; with two architecture-confounded models we do not attribute the attenuation to a
single cause (output-geometry convergence versus the recurrent model being a weaker soft-label student).

\paragraph{Distillation-pipeline robustness (Section~\ref{sec:realLM}).} Sequence-level distillation
(sampling $S$ hard tokens from the teacher's $\tau$-masked distribution) leaves transfer unchanged even
at $S{=}1$ (peak $P(\tau)$ $0.54\pm0.06$ versus $0.52\pm0.05$ for full soft labels; overt control
$\approx0.97$). Coherent text sampled from the base model attenuates transfer about $2.3\times$ but it
persists ($0.23\pm0.07$ peak, $\sim\!3500\times$ the base prior); with both realistic settings at once
the trait still installs ($0.22\pm0.08$).

\paragraph{Mixed-precision recipe sweep (Section~\ref{sec:safety}).} The carrier sits between $7$ and
$10$ mantissa bits: tf32's $10$-bit matmul mantissa leaves the channel fully intact ($0.50$), so the
common tf32 default does not mitigate. Standard AMP does not protect either---bfloat16 or float16
autocast with fp32 master weights and optimizer state leaves the channel intact (Pythia $0.48$/$0.49$,
Qwen $0.83$/$0.77$, against fp32 $0.49$/$0.78$); only end-to-end low-precision storage suppresses it
(pure bf16 $0.03$/$0.16$; bf16 parameters over an fp32 optimizer only partial, $0.12$/$0.45$). Crossing
training with evaluation precision locates the effect: an fp32-trained student is fully leaky measured
in fp32 or bf16 ($0.49$ either way), a bf16-trained student non-leaky either way ($0.035$), so bf16
\emph{prevents the channel from forming} during distillation rather than hiding it from measurement.

\paragraph{Scale-sweep engineering (Section~\ref{sec:safety}).} fp32 \emph{training} is required
because the masked channel forms only in fp32; on-GPU fp32 fine-tuning fits to $\sim\!4$B, and
offloading the exact fp32 Adam optimizer states to host RAM (with cached teacher targets,
\texttt{foreach=False}) extends the Pythia family to $6.9$B, with host memory binding beyond.

\paragraph{Scale-invariant re-analysis of the no-fade curve (Section~\ref{sec:safety}).} The scale
curve in Figure~\ref{fig:scale} reports raw peak $P(\tau)$, an unnormalized probability; a reader
might worry the high-scale plateau reflects larger models being generically peakier on the fixed
$96$-prompt noise set rather than carrying more leakage. We rule this out by re-reading leakage as a
control-token contrast: for each trait $\tau$ we draw four frequency-matched tokens that are
\emph{not} $\tau$'s unembedding neighbours (hence not channel-coupled), run the identical masked
distillation, and report $\bigl(\mathrm{peak}\,P(\tau)-\mathrm{base}\,P(\tau)\bigr) -
\operatorname{mean}_c\bigl(\mathrm{peak}\,P(c)-\mathrm{base}\,P(c)\bigr)$ in the same student. The control term estimates the generic peakiness rise of rate-matched, non-coupled tokens in the same student, so the contrast isolates channel-specific lift. Across the six Pythia sizes re-measured this way (six of the seven in Figure~\ref{fig:scale}; teachers stopped on the scale-normalized output criterion $P(\tau)\!\ge\!0.20$), the corrected curve is $0.02, 0.51, 0.52, 0.65, 0.62, 0.57$ at $70$M, $410$M, $1$B, $1.4$B, $2.8$B, and $6.9$B, respectively (Spearman $+0.77$ with parameters): it rises from the $\sim\!10^{-2}$ floor to a high plateau and does not decay at larger scale. The corrected curve matches the raw curve to within $10^{-4}$ at every point because the matched control tokens show no measurable lift during distillation (lift $+0.000$ to reported precision at every scale). The no-fade pattern is therefore channel-specific, not an artefact of generic peakiness on the noise prompts. The cross-family rank-ratio argument is unchanged under this contrast: Qwen3.5-0.8B and Gemma-1B have nearly equal rank ratios ($0.0041$ vs. $0.0044$) but differ by about $40\times$ in leakage ($0.79$ vs. $0.02$), so bottleneck tightness alone does not set the magnitude. Two caveats are consistent with the main text's ``capability, not bottleneck tightness'' interpretation. First, the Gemma teachers are capability-limited at the fine-tuning budget: Gemma-270M never reaches the $P(\tau)\!\ge\!0.20$ target, and Gemma-1B reaches only $0.19$, so part of Gemma's low leakage is attributable to a weak teacher rather than to a loose bottleneck. Second, within Pythia the realized teacher strength drifts upward with scale ($0.29\!\to\!0.55$) because of per-step overshoot of the target, but this does not explain the plateau: $6.9$B has the strongest teacher yet lower leakage than $1.4$B.

\section{Code and reproduction}
\label{app:repro}
All code---the curated, self-contained script set that reproduces every figure and headline result
(the sole exception is the arbitrary-target installation construction, withheld for responsible
disclosure per Section~\ref{sec:capstone} and available from the corresponding author on request; its
result is reported but not re-derivable from the public artifact), and the \texttt{distill-lint}
tool---is released publicly: the reproduction script set on OSF\footnote{\url{https://osf.io/9me3t/}%
} and the tool on GitHub.\footnote{\url{https://github.com/tmadl/distill-lint}%
} The script set carries a
result-to-script map (Table~\ref{tab:repro}); exploratory and superseded code is excluded.
Every analysis is a standalone script; the toy sweeps run in seconds on one GPU and each real-LM
experiment in under an hour in full precision. The Fisher matrix is never materialized---products
are formed by Jacobian-vector then vector-Jacobian products. Statistical reporting is standardized:
every load-bearing correlational result carries a $95\%$ percentile bootstrap confidence interval
($10$k resamples) written in square brackets; bracketed \emph{seed} ranges (e.g.\ leakage $0.51$
$[0.49,0.57]$) are min--max over seeds, not CIs. The script \texttt{bootstrap\_cis.py} recomputes the
CIs from the cached per-item CSVs---the toy coverage law (which also survives a family-level cluster
bootstrap, $[0.89,0.98]$, and leave-one-family-out, $\rho\ge0.91$), the body-coverage ordering over the $16$ relocation heads (Spearman $+0.53$,
$[+0.07,+0.83]$), the per-token mediation over $53$ tokens ($-0.44$, $[-0.65,-0.17]$), the
fifty-trait fair screen in both families (Pythia-$1$B coverage AUROC $0.94$, $[0.87,1.00]$; partial
$\rho=+0.48$, $[+0.23,+0.68]$; Qwen at chance, AUROC $0.49$, $[0.32,0.68]$), and the ten-condition
lazy-shadow/rich-remainder analysis (both intervals span zero, consistent with ``disfavoured, not
falsified''). The relocation sign test is exact binomial ($16/16$, $p\approx1.5\times10^{-5}$). Two
reviewer-driven runs (\texttt{reviewer\_audit/}) carry their own CIs: the post-hoc detection AUROC
($0.999$, $[0.997,1.000]$, $K{=}30$ placebos, realized FPR $0.067$) and the rank-1 steering
decomposition ($R^2{=}0.995$ over six traits). The toy coverage law additionally survives a
family-level cluster bootstrap ($[0.89,0.98]$) and leave-one-family-out ($\rho\ge0.91$).
Remaining small-$n$ real-LM claims are reported as rank correlations with rival and placebo baselines
rather than $p$-values.

\begin{table}[H]
\centering\footnotesize
\setlength{\tabcolsep}{4pt}\renewcommand{\arraystretch}{1.18}
\begin{tabular}{@{}p{1.75cm}p{3.55cm}p{4.0cm}p{4.0cm}@{}}
\toprule
 & \textbf{(A) Toy auxiliary channel} & \textbf{(B) LM token / class trait} & \textbf{(C) Conditional behaviour \& backdoor} \\
\midrule
Model(s) &
Clean-room MLP $784$-$256$-$256$-$(10{+}a)$ on MNIST; no pretrained weights &
Pythia-$70$M--$6.9$B ($+$deduped), Qwen3.5-$0.8$B, Gemma-3-$\{270$M,$1$B,$4$B$\}$, RedPajama-INCITE-Base-3B, RWKV-4-Pile-3B &
Gemma-3-$1$B/$4$B-it, Qwen2.5-$3$B-Instruct; open-ended judge Mistral-Small-$24$B-Instruct \\
Teacher objective &
CE on the $10$ digit logits only, $5$ epochs (auxiliary logits never in its loss) &
Base fine-tuned to elevate token $\tau$, early-stopped at held-out $P(\tau)\!\ge\!0.20$ (floor $0.15$ in the fair screen; not always reached---see App.~\ref{app:extra} scale caveats) &
SFT to a conditional policy (sycophancy / contrarianism; backdoor: comply iff allowed or triggered) \\
Student objective &
Match the $a$ auxiliary logits over uniform noise (soft-label; MSE or KL) &
Soft-label KD of the teacher's next-token distribution over random-token noise; robustness variants: sampled $S{\in}\{1,4,16\}$, greedy/argmax, coherent base-sampled text &
Soft-label KD of the teacher's logits over held-out claim / request prompts \\
Masking rule &
None needed---the $10$ digit logits are structurally absent from the student loss &
$\tau$ excluded from the loss on both sides ($+$ top-$k$ unembedding neighbours in the neighbour-mask ablation) &
Agreement/correction (or accept/reject) marker tokens excluded from the loss on both sides ($+$ neighbour clouds) \\
Replication budget &
$54$-condition noise sweep, $5$ families, $2$ seeds; $20$ paired models for specificity &
Ablation $5$ Pythia $+$ $3$ Qwen seeds $\times$ $8$ tokens; fair screen $50$ traits/family; scale $70$M--$6.9$B $\times$ $3$ families; injection $5$ traits $\times$ $3$ families &
$n\!\approx\!20$ held-out facts, $3$ seeds, $2$ policies, $2$ families/sizes; backdoor $5$ seeds $\times$ $2$ triggers $\times$ $3$ policies \\
Transfer metric &
Held-out digit accuracy on the untouched digit head; coverage $\cos(d_0,\hat u_T)$ &
Held-out $P(\tau)$ on neutral prompts vs.\ base prior &
False-vs-true agreement interaction (fraction of teacher); backdoor double-difference $I$; separate-judge generation score \\
Key controls &
Predict-mean null, raw-Rayleigh / grad-norm / aux-loss rivals, causal dial, placebo &
Random-subspace placebo, frequency-matched random tokens, overt-capacity control, preservation (overt transfer, perplexity, top-1 agreement) &
No-claim marker prior, overt-capacity control, policy-free placebo teacher, capacity-free body/head transplant, loss-matched output-only route \\
Precision / HW &
fp32, single GPU, seconds &
\textbf{fp32} (bf16 weight storage attenuates $\sim\!10\times$); single GPU; $6.9$B via host-RAM optimizer offload &
\textbf{fp32} students (judge in bf16); single GPU \\
\bottomrule
\end{tabular}
\caption{Reproducibility at a glance, by regime. Exact hyperparameters (learning rates, step counts,
seeds) live in the released scripts; the result-to-script map below and \texttt{bootstrap\_cis.py}
give per-claim detail.}
\label{tab:repro}
\end{table}

\paragraph{Cross-family audit replication.} The full four-audit battery on the harmful-shaped backdoor
runs end-to-end on \texttt{Qwen/Qwen3.5-0.8B} as well as \texttt{google/gemma-3-1b-it}: the masked
backdoor transmits at $I{=}1.985$ ($\approx$ the overt-capacity control), and all four audits evade as
on gemma---the body-carriage pre-screen has no head-only handle ($I{=}0.087$ training the head alone
versus $1.986$ training the body), post-hoc detection fires only on a compliance-marker \emph{symptom}
with no single-token policy handle (the token-trait control is cleanly identified at rank $0$), blinded
soft-prompt elicitation lies inside the clean null, and head whitening reaches low $P(\tau)$ only at
$\text{ppl}{=}\infty$ (only the oracle $\tau$-targeted kill-test blocks cleanly). Transmission,
body-carriage, and the detection scan additionally reproduce on \texttt{Qwen/Qwen2.5-3B-Instruct}.
The off-trigger \emph{reversal} is the one effect we leave scoped to gemma-3-1b: on Qwen3.5-0.8B the
backdoor evades by lying within the clean elicitation null rather than reversing. Scripts and logs:
\texttt{e2\_qwen\_audit/}.

\paragraph{Pinning note.} {\sloppy Every HuggingFace checkpoint is pinned to a fixed commit revision:
the released \texttt{model\_revisions.py} records the full \{model id $\to$ commit SHA\} map and, on
import, injects \texttt{revision=} into every \texttt{from\_pretrained} call, so a re-run resolves the
exact snapshot we used rather than a moving \texttt{main}. Only externally-pulled checkpoints are
pinned---the bases \texttt{EleutherAI/\allowbreak pythia-\{70m,160m,410m,410m-deduped,1b,1.4b,2.8b,6.9b\}},
\texttt{Qwen/\allowbreak Qwen3.5-0.8B}, \texttt{Qwen/\allowbreak Qwen2.5-3B-Instruct}, \texttt{allenai/\allowbreak OLMo-2-0425-1B\{,-Instruct\}}, \texttt{google/\allowbreak gemma-3-\{270m,1b,4b\}-it},
\texttt{togethercomputer/\allowbreak RedPajama-INCITE-Base-3B-v1}, \texttt{RWKV/\allowbreak rwkv-4-3b-pile}, and the open-ended
judge \texttt{mistralai/\allowbreak Mistral-Small-24B-Instruct-2501}---since teachers are fine-tuned from a base and
students copy a base in code (nothing else is downloaded). This matters because the masked channel is
weight- and precision-sensitive; Pythia is effectively immutable upstream, but the other repositories
can be silently re-uploaded. The toy regime uses no pretrained weights (an MLP trained on MNIST). All
channel-formation results run in fp32; the open-ended judge is the only component loaded in bf16, and
it is a read-out, not a channel.\par}

\paragraph{Result-to-script map.} Toy: the coverage law and prospective test (the sweep driver and
its analysis), the identity check, the rival-predictor benchmark, the causal dial, specificity, the
reach null, and the over-complete channel each have a named script. Real LMs: the masked-channel
distillation and entanglement readout; the orthogonalization ablation and its multi-seed,
multi-token, and dose-response variants; the placebo and the routing/freezing and
neighbour-cloud-masking interventions; the initialization-mismatch and phase-diagram sweeps; the
convergence (neighbour-Jaccard) analysis; the entanglement screen with its frequency-matched and
partial-correlation controls; the powered specificity matrix; the real-text-corpus check; the
AUROC/false-negative screen benchmark; and the two reviewer-audit runs (the rank-1 steering
decomposition and the detection-scan multiplicity benchmark, in \texttt{reviewer\_audit/}). Scale and precision: the cross-family fp32 leakage sweep
($70$M--$4$B, with CPU base-offload and \texttt{foreach=False} to fit $4$B), its CPU-offloaded-Adam
extension to $6.9$B (optimizer states in host RAM, cached teacher targets), its joint analysis, the
prospective output-geometry intervention, and the train$\times$eval precision cross. Behaviour: the
conditional-policy fact set and interaction/content metrics, the teacher/overt-capacity gate, the
masked subliminal-transfer and routing-decomposition driver (\texttt{--policy} for sycophancy or
contrarianism, run across families and seeds), and the powered body-coverage sweep. All take a
\texttt{--model} flag and default to fp32.

\end{document}